\documentclass[preprint,3p,times,authoryear]{elsarticle}

\usepackage{amssymb}
\usepackage{amsthm}
\usepackage{subfig}
\usepackage{caption}
\usepackage{hyperref}
\usepackage{microtype}
\usepackage{graphicx}
\usepackage{booktabs} 
\usepackage[table]{xcolor}

\usepackage{amsmath}
\usepackage{amssymb}
\usepackage{mathtools}
\usepackage{amsthm}
\usepackage{makecell}
\usepackage{multirow}
\usepackage{csquotes}

\usepackage{lineno}

\journal{GIScience \& Remote Sensing}
\begin{document}

\begin{frontmatter}



\title{LMSeg: An end-to-end geometric message-passing network on barycentric dual graphs for large-scale landscape mesh segmentation}

\author[inst1]{Zexian Huang\corref{cor1}}
\cortext[cor1]{Corresponding author}
\ead{zexianh@student.unimelb.edu.au}

\author[inst1]{Kourosh Khoshelham}
\ead{k.khoshelham@unimelb.edu.au}

\author[inst1]{Martin Tomko}
\ead{tomkom@unimelb.edu.au}

\affiliation[inst1]{organization={The University of Melbourne},
            city={Parkville},
            postcode={3010}, 
            state={Victoria},
            country={Australia}}

\begin{abstract}
Semantic segmentation of large-scale 3D landscape meshes is critical for geospatial analysis in complex environments, yet existing approaches face persistent challenges of scalability, end-to-end trainability, and accurate segmentation of small and irregular objects. To address these issues, we introduce the BudjBim Wall (BBW) dataset, a large-scale annotated mesh dataset derived from high-resolution LiDAR scans of the UNESCO World Heritage-listed Budj Bim cultural landscape in Victoria, Australia. The BBW dataset captures historic dry-stone wall structures that are difficult to detect under vegetation occlusion, supporting research in underrepresented cultural heritage contexts. Building on this dataset, we propose LMSeg, a deep graph message-passing network for semantic segmentation of large-scale meshes. LMSeg employs a barycentric dual graph representation of mesh faces and introduces the Geometry Aggregation+ (GA+) module, a learnable softmax-based operator that adaptively combines neighborhood features and captures high-frequency geometric variations. A hierarchical–local dual pooling integrates hierarchical and local geometric aggregation to balance global context with fine-detail preservation. Experiments on three large-scale benchmarks (SUM, H3D, and BBW) show that LMSeg achieves 75.1\% mIoU on SUM, 78.4\% O.A. on H3D, and 62.4\% mIoU on BBW, using only 2.4M lightweight parameters and outperforming strong point- and graph-based baselines. In particular, LMSeg demonstrates accurate segmentation across both urban and natural scenes—capturing small-object classes such as vehicles and high vegetation in complex city environments, while also reliably detecting dry-stone walls in dense, occluded rural landscapes. Together, the BBW dataset and LMSeg provide a practical and extensible method for advancing 3D mesh segmentation in cultural heritage, environmental monitoring, and urban applications.
\end{abstract}




\end{frontmatter}



\section{Introduction}
\label{introduction}

Recent advancements in terrain reconstruction and data acquisition have enabled automated modeling of large-scale landscapes. Such landscapes are typically represented as 2.5D rasters (e.g., digital elevation models) or as textured 3D triangle meshes derived from Triangular Irregular Networks (TINs) or Multi-View Stereo (MVS) reconstructions. Among these, 3D meshes preserve geometric structures (via triangle connectivity), semantic surface attributes (e.g., RGB texture, face normals), and topological adjacency, offering richer information than rasters. However, semantic segmentation of 3D meshes remains less developed than raster-based approaches. Raster methods benefit from regular grid structures that align well with convolutional networks and Vision Transformers, whereas meshes are irregular, non-Euclidean, and computationally demanding. Many existing mesh segmentation techniques still depend on handcrafted features or multi-stage pipelines, limiting their scalability and generalizability.

Mesh learning faces three persistent challenges: (1) \textit{Limited scalability}: most existing methods are designed for small, preprocessed, and clean mesh datasets, which restricts their applicability to real-world large-scale scenarios. (2) \textit{Lack of end-to-end trainability}: large-scale mesh processing often depends on non-differentiable, multi-stage heuristics or handcrafted descriptors, limiting the trainability and adaptability of segmentation models. (3) \textit{Difficulty in segmenting fine-grained and irregular objects}: small or topologically complex structures (e.g., walls, vehicles, boats) remain challenging, as geometric noise and severe class imbalance obscure boundaries and reduce accuracy. To address these issues and support research in underrepresented cultural heritage contexts, we introduce the BudjBim Wall (BBW) dataset, a large-scale annotated mesh dataset for segmenting dry-stone wall structures in complex vegetated terrain. Derived from high-resolution LiDAR scans of the UNESCO World Heritage-listed Budj Bim cultural landscape in Victoria, Australia, the BBW dataset captures heritage features of ecological and cultural significance that are often obscured by vegetation and difficult to detect with conventional mapping methods.

Alongside this dataset, we propose the \textbf{L}andscape \textbf{M}esh \textbf{Seg}mentation Network (LMSeg), a deep graph message-passing network tailored for large-scale landscape meshes. LMSeg employs the barycentric dual graph of the mesh as its input representation, enabling joint learning on both mesh features and topological structure. Each node encodes geometric attributes (e.g., normals, coordinates) and semantic attributes (e.g., RGB), while edges capture explicit face adjacency. This barycentric dual graph representation avoids reliance on handcrafted features \citep{rouhani2017semantic,gao2021sum,gao2023pssnet} and inherently supports end-to-end trainability through graph message passing. A core component of LMSeg is the \textbf{Geometry Aggregation+} (GA+) module, which introduces a parameterized softmax-based aggregation operator that generalizes traditional GNN aggregators (\texttt{mean}, \texttt{sum}, \texttt{max}) and adaptively combines graph neighborhood features. Combined with trigonometric positional encoding, GA+ effectively captures high-frequency geometric variations, thereby improving segmentation of small and irregular features. Furthermore, LMSeg integrates hierarchical (HGA+) and local (LGA+) pooling to balance global contextual reasoning with boundary refinement, ensuring robustness across scales. The main contributions of this work are as follows:
\begin{itemize}
    \item \textbf{BudjBim Wall (BBW) Dataset}: We introduce a large-scale annotated 3D mesh dataset for detecting historic dry-stone walls under vegetation occlusion, supporting segmentation research in underrepresented cultural heritage landscapes.
    \item \textbf{Barycentric Dual Graph Representation}: We formulate a face-centric dual graph that enables per-face feature learning, improves message-passing consistency, supports end-to-end graph message-passing trainability, and eliminates reliance on handcrafted descriptors.
    \item \textbf{Geometry Aggregation+ (GA+) Module}: We propose a learnable softmax-based aggregation operator, coupled with trigonometric positional encoding, that captures high-frequency geometric variations and adaptively combines neighborhood features.
    \item \textbf{Hierarchical \& local dual pooling}: We integrate hierarchical (HGA+) and local (LGA+) geometric aggregation to jointly enhance multi-scale contextual reasoning and preserve fine structural details.
\end{itemize}

\section{Background}  
\subsection{Surface Representations and Segmentation}  
The segmentation of landscape surfaces, including those with or without explicit 3D objects, has long been a central topic in the spatial sciences. Extensive studies have addressed semantic segmentation of real-world, large-scale landscapes at increasingly fine resolutions \citep{blaschke2010object}. Three major representations exist for such landscapes: (1) irregularly sampled point clouds capturing 3D coordinates of surface points, (2) regular grid-based raster representations of aggregated elevation values per cell, and (3) mesh representations of Triangular Irregular Networks (TINs) \citep{peucker1976digital}. These representations are often connected through transformation pipelines. For example, lidar-derived unstructured point clouds may be interpolated into raster grids for 2D image-based analysis, or triangulated into TIN meshes to explicitly model surface connectivity. Compared to rasters, meshes preserve the original 3D geometry of sampled points while also encoding topological relationships between facets, including connectivity and orientation (normals) (see Section~\ref{sec:dual}).  

Different segmentation approaches have been developed for these representations. In raster-based segmentation, deep learning methods typically employ convolutional neural networks (CNNs) \citep{he2016deep,girshick2014rich} to extract semantic features and classify pixels. Recent advances, such as ViT-based models \citep{wang2022unetformer,dosovitskiy2020image}, have achieved state-of-the-art performance in remote sensing imagery. However, purely 2D raster methods are inherently limited by occlusion, loss of 3D context, and the absence of explicit surface topology modeling.  

Recent multimodal remote sensing semantic image segmentation (MRSISS) extends single-modal raster methods by integrating complementary modalities (e.g., optical and synthetic aperture radar). \citet{ye2025tuple} proposed TMCNet, a contrastive learning framework that learns shared and modality-specific representations, followed by a lightweight segmentation network (MSSNet). \citet{zhou2025advancing} introduced FusionMatch, a semi-supervised framework that integrates near-infrared imagery and pan-sharpening fusion to expand the perturbation space. In dense urban prediction, MDNet \citep{zhou2024mdnet} incorporates a Mamba-based fusion module for long-range dependencies and a diffusion self-distillation strategy for feature refinement. While effective for multimodal imagery, these approaches remain tied to raster inputs and cannot directly exploit mesh topology.  

For 3D point cloud segmentation, point-based methods \citep{qi2017pointnet,qi2017pointnet++,tailor2021towards,zhao2021point,hu2020randla,thomas2019kpconv} and graph-based methods \citep{wang2019dynamic,li2019deepgcns,landrieu2018large} demonstrate the importance of local neighborhood learning. DeepGCN \citep{li2019deepgcns} further advanced this direction by introducing residual connections and normalization strategies to enable training very deep GNNs, mitigating oversmoothing and gradient vanishing. Nonetheless, recent survey \citep{adam2023deep} highlights that both point- and graph-based methods underperform in landscape-scale segmentation due to challenges in learning features on large, complex mesh surfaces and structures. Multimodal extensions attempt to address this limitation: MFFNet \citep{ren2024mffnet}, for example, projects raw point clouds into RGB and frequency images for efficient semantic extraction with 2D convolutions, then fuses back into the 3D domain using local attention. Although effective, such approaches do not explicitly model mesh topology or support end-to-end mesh learning.  

Across raster, point, and multimodal approaches, a common limitation persists: the absence of direct, end-to-end learning on mesh structures that encode both geometry and topology. Graph U-Net \citep{gao2019graph} pioneered hierarchical pooling and unpooling for graphs, showing the potential of multi-scale propagation, but was primarily evaluated on non-spatial graphs (i.e., citation and social graphs) rather than geometric meshes, limiting transferability to geospatial segmentation tasks. \textbf{LMSeg} addresses this gap by converting meshes into barycentric dual graphs and applying a hierarchical–local graph learning architecture that jointly captures global structure and fine-scale geometric detail, enabling efficient and robust large-scale mesh segmentation.  

\subsection{Mesh-based Surface Segmentation}  
In contrast to raster-, point-, and graph-based approaches, mesh-based methods directly process TINs or multi-view stereo (MVS) meshes \citep{gao2021sum,rouhani2017semantic}. Early pipelines, such as \citet{rouhani2017semantic}, segmented textured meshes into superfacets using region growing, then extracted geometric and photometric descriptors for classification via decision trees and refinement with a Markov Random Field. \citet{gao2021sum} extended this design with semi-automatic annotation and a random forest classifier for large-scale urban meshes. While achieving stable results, these pipelines depended heavily on over-segmentation and handcrafted features, limiting scalability and end-to-end optimization.  

Learning-based mesh methods advanced toward intrinsic mesh processing. MeshCNN \citep{hanocka2019meshcnn} applies convolutions on mesh edges, while MeshNet \citep{feng2019meshnet} and MeshNet++ \citep{singh2021meshnet++} represent faces using spatial and structural descriptors. Graph-based methods such as DenseGCN \citep{tang2021dense} treat meshes as densely connected graphs, while LGCPNet \citep{guan2021lgcpnet} constructs dual local graphs on barycentric points. Hierarchical pooling approaches, inspired by Graph U-Net \citep{gao2019graph}, also highlight the importance of multi-scale learning, though their pooling is not explicitly geometry-aware. More recently, self-attention mechanisms have been explored \citep{li2022laplacian,basu2022equivariant} to capture long-range dependencies in mesh classification and segmentation. However, most of these models are primarily tested on small, clean synthetic datasets \citep{bronstein2011shape,wu20153d}, limiting their demonstrated applicability to noisy, incomplete, or large-scale landscape meshes.

\subsection{Learning Models for Urban Meshes}  
Several works have specifically targeted real-world urban meshes. \citet{tutzauer2019semantic} projected mesh face barycenters into point clouds and trained multi-branch CNNs on geometric and radiometric features. \citet{laupheimer2020association} fused lidar point clouds with textured meshes and applied PointNet++ for semantic segmentation. \citet{tang2022deep} abstracted meshes to barycentric point clouds and employed a point-based Transformer \citep{zhao2021point} for face classification, though at high computational cost. \citet{yang2023surface} represented meshes as barycentric dual graphs with hierarchical neighborhoods and texture convolutions, but relied on feature engineering and multi-stage heuristics that limited end-to-end optimization. PSSNet \citep{gao2023pssnet} introduced a two-stage design based on superfacet grouping and edge dependency extraction, further disrupting joint training.  

These urban mesh approaches illustrate steady progress toward graph-based processing of surface topology, yet they often depend on intermediate representations (e.g., barycentric points, superfacets) or handcrafted descriptors. Building on this trend, \citet{liu2024umeshsegnet} demonstrated the benefits of mesh-native attention mechanisms for UAV photogrammetric data, achieving improved accuracy on urban mesh datasets. Nevertheless, such methods still struggle to balance global context with fine-grained detail and face challenges in scaling to irregular, noisy geospatial meshes.  

To address these limitations, the proposed \textbf{LMSeg} operates directly on the barycentric dual graph and unifies hierarchical (HGA+) and local (LGA+) aggregation. This architecture balances global structural context with fine-scale boundary refinement within a single, end-to-end trainable graph message-passing framework. As a result, LMSeg demonstrates robustness on noisy, irregular large-scale meshes and achieves more accurate segmentation of small and complex landscape features across multiple semantic mesh benchmarks.

\section{Methodology}

\begin{figure*}[!tbp]
\centering
\includegraphics[width=1\textwidth]{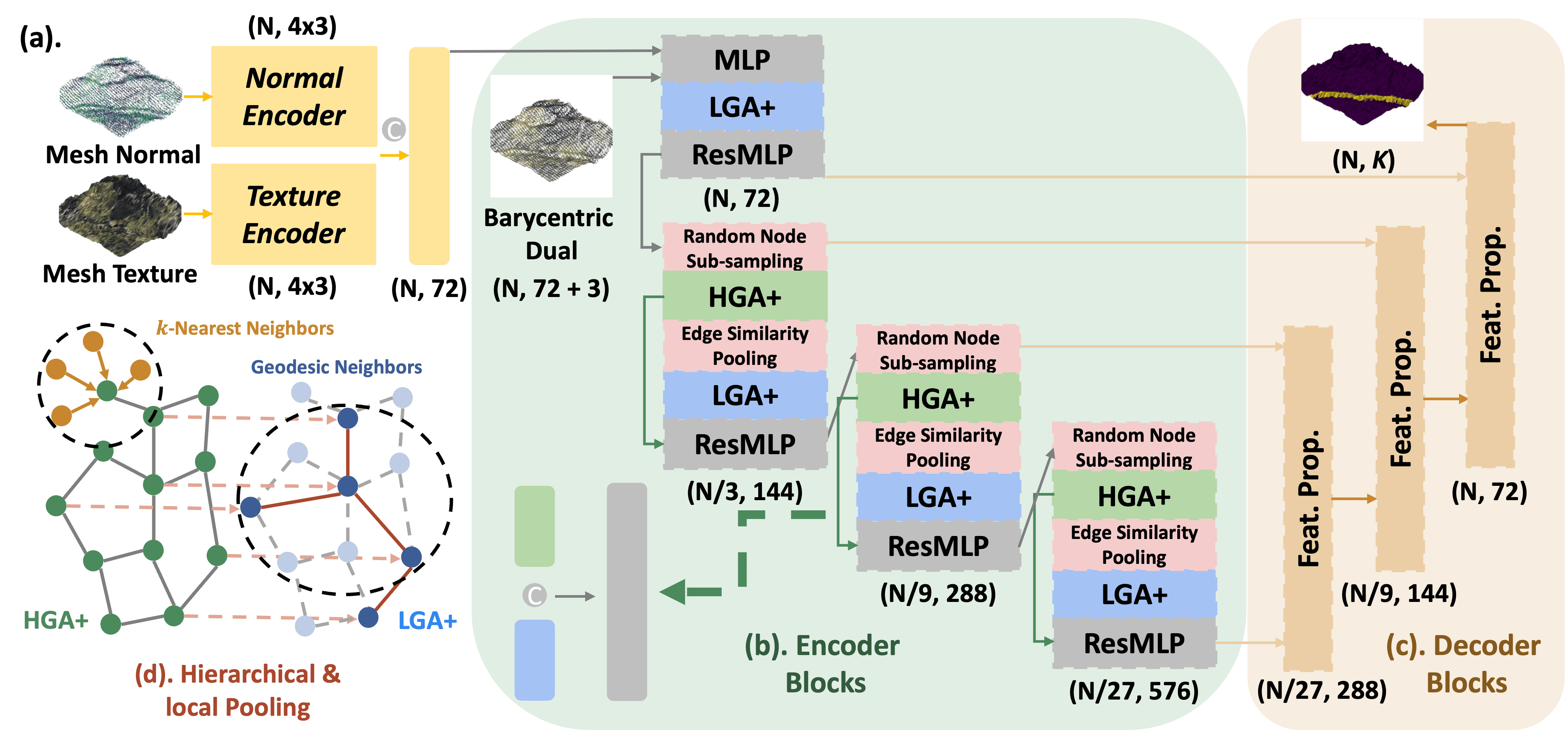}
\caption{Overall architecture of LMSeg (Large-scale Mesh Segmentation Network).  
(a) The input triangular mesh is converted into a \textbf{barycentric dual graph}, where each node corresponds to a triangular face and edges capture 1-ring face adjacency. Node features consist of RGB values and surface normals from both faces and vertices, each processed by a \textbf{Mesh Feature Encoder} into a shared latent space.  
(b) The \textbf{encoder} alternates two complementary modules: \textbf{HGA+} (Hierarchical Geometry Aggregation+) and \textbf{LGA+} (Local Geometry Aggregation+), separated by random node sub-sampling and edge similarity pooling. \textbf{HGA+} operates on a $k$-nearest neighbor graph linking downsampled nodes to their neighbors in the full-resolution mesh, capturing hierarchical, long-range, and multi-scale structural context. \textbf{LGA+} operates on locally pooled geodesic neighborhoods, refining features using high-frequency, fine-scale variations to preserve geometric detail. Outputs from HGA+ and LGA+ are concatenated and refined by a residual multilayer perceptron (\textbf{ResMLP}). Each stage reduces node count while increasing feature dimensionality.  
(c) The \textbf{decoder} progressively upsamples features back to the original resolution using inverse-distance interpolation, skip connections, and MLP refinement, producing dense per-face semantic predictions.  
(d) Illustration of the hierarchical and local pooling strategy: random node sub-sampling reduces graph size efficiently, while edge similarity pooling restores meaningful local neighborhoods. HGA+ operates on hierarchical neighborhoods defined in the original geometry, and LGA+ operates on local geodesic neighborhoods, ensuring explicit multi-scale feature fusion.  
Here, \texttt{N} denotes the number of nodes in the barycentric dual graph, \texttt{D} the feature dimensionality, and \texttt{K} the number of semantic classes.}
\label{fig:architecture}
\end{figure*}

\label{approach}
In this section, we introduce the building blocks of LMSeg, a deep message-passing network proposed for effective and accurate large-scale landscape mesh segmentation tasks. The overall LMSeg architecture follows an encoder–decoder design (Fig.~\ref{fig:architecture}). Section~\ref{sec:dual} provides the formal definition of landscape meshes and the barycentric dual graph input used by LMSeg. The overall network architecture is described in Section~\ref{sec:archi}. Section~\ref{sec:gap} presents the \textbf{Geometry Aggregation+} (GA+) module, which performs adaptive, geometry-aware aggregation to learn expressive local geometric latent embeddings on the barycentric dual graph $\mathcal{G}(\mathcal{M})$. Section~\ref{sec:pooling} discusses the hierarchical node sub-sampling and local edge pooling strategies used in LMSeg.

\subsection{Triangular Landscape Mesh and its Barycentric Dual Graph}
\label{sec:dual}
We define \enquote{landscape meshes} as triangular meshes representing both urban and natural landscapes, including 2.5D TIN models \citep{peucker1976digital} and full 3D MVS meshes \citep{rouhani2017semantic, gao2021sum}. These meshes capture the geometric complexity of terrain, vegetation, and built structures. While they may include radiometric (texture) information—particularly in the case of MVS-derived models—our definition does not require the presence of texture.

Consider a landscape surface $\mathcal{S}$ represented as a triangular mesh $\mathcal{M} = (\mathcal{V}, \mathcal{E}, \mathcal{F})$, where $\mathcal{V} \in \mathbb{R}^{N \times 3}$ is a set of unordered points with spatial coordinates on $\mathcal{S}$, $\mathcal{E} \subseteq \mathcal{V} \times \mathcal{V}$ denotes a set of undirected edges, and $\mathcal{F} = \{(i, j, k) \in \mathcal{V} \}$ is a set of ordered triplets constituting the triangular faces. Together, $\mathcal{V}$, $\mathcal{E}$, and $\mathcal{F}$ form a 3D mesh capturing the geometry of the surface in real-world environments.

The triangle mesh $\mathcal{M}$ is converted into its topological dual, the barycentric graph $\mathcal{G}(\mathcal{M})$. Nodes correspond to barycenters of triangular faces, edges capture face adjacency, and node features $\mathcal{X} \in \mathbb{R}^{N \times C}$ store mesh attributes such as normals and texture. Compared to point cloud or vertex-based representations, the barycentric dual graph offers a structured, face-centric formulation that enables direct per-face supervision, efficient aggregation of geometric and textural information, and, critically, avoids costly resampling on unevenly distributed large-scale meshes. This makes the barycentric dual graph particularly suitable for landscape-scale segmentation tasks where meshes often exhibit irregular density and complex topologies.

\subsection{Network Architecture}
\label{sec:archi}

\paragraph{Inputs (Fig.~\ref{fig:architecture}(a)):} LMSeg takes as input the barycentric dual graph $\mathcal{G}(\mathcal{M})$, where each node corresponds to a mesh face and edges represent 1-ring face adjacency. Node features include RGB values and normal vectors extracted from both the face and its three associated vertices (four $3$D vectors each for RGB and normals, forming two $4\times3$ tensors). Face normals encode the average surface orientation of each triangle, while vertex normals capture local curvature and smoothness, together providing complementary geometric cues for distinguishing between planar, curved, and irregular surfaces. Similarly, face RGB values represent the mean texture color across a triangle, while vertex RGB values preserve fine-scale texture variation from the underlying image or photogrammetric reconstruction. These combined geometric and photometric attributes provide rich, complementary information for separating semantic classes with similar shapes but different appearances. A \texttt{MeshFeatureEncoder} projects these features into a shared latent embedding via a shared MLP, then aggregates them using a temperature-scaled softmax operator (Eq.~\ref{eq:aggr_softmax}). The resulting per-face embedding integrates both textural appearance and surface geometry, and is concatenated with barycentric coordinates to form the final feature vector $\mathbf{x} \in \mathbb{R}^{N \times D}$, which is passed to the first GA+ module.

\paragraph{Encoder (Fig.~\ref{fig:architecture}(b)):} Each encoder stage applies: random node sub-sampling $\rightarrow$ HGA+ $\rightarrow$ edge similarity pooling $\rightarrow$ LGA+ $\rightarrow$ residual MLP. HGA+ operates on a $k$-nearest neighbor graph between full-resolution and downsampled nodes, capturing hierarchical, multi-scale context; LGA+ works on the locally reconstructed pooled graph, preserving fine-grained geometric detail. Alternating HGA+ and LGA+ modules allows the network to progressively fuse global structural cues with localized high-frequency features—critical for segmenting small or irregular objects—while random sampling and pooling keep memory and runtime costs low, supporting efficient large-scale inference. Outputs from HGA+ and LGA+ are concatenated and passed through a ResMLP. Each stage reduces node count by $\frac{1}{3}$ and doubles feature dimensionality.

\paragraph{Decoder (Fig.~\ref{fig:architecture}(c)):} The decoder mirrors the encoder, upsampling features back to input resolution using inverse-distance interpolation over three adjacent neighbors, followed by skip connections and MLP refinement.

\subsection{Geometry Aggregation+ Module (GA+)}
\label{sec:gap}
The Geometry Aggregation+ (GA+) module is the core feature learning block of LMSeg, designed for geometry-aware representation learning on the barycentric dual graph $\mathcal{G}(\mathcal{M})$. GA+ addresses the limitations of fixed aggregation functions (\texttt{sum}, \texttt{mean}, \texttt{max}) in conventional graph message passing, which often fail to adapt to the varying geometric structures present in large-scale 3D meshes. In particular, small or irregular objects occupy only a few mesh faces and require feature aggregation that can capture fine-grained, high-frequency geometric patterns while preserving broader structural context.

\begin{figure}[!tbp]
\centering
\includegraphics[width=0.5\textwidth]{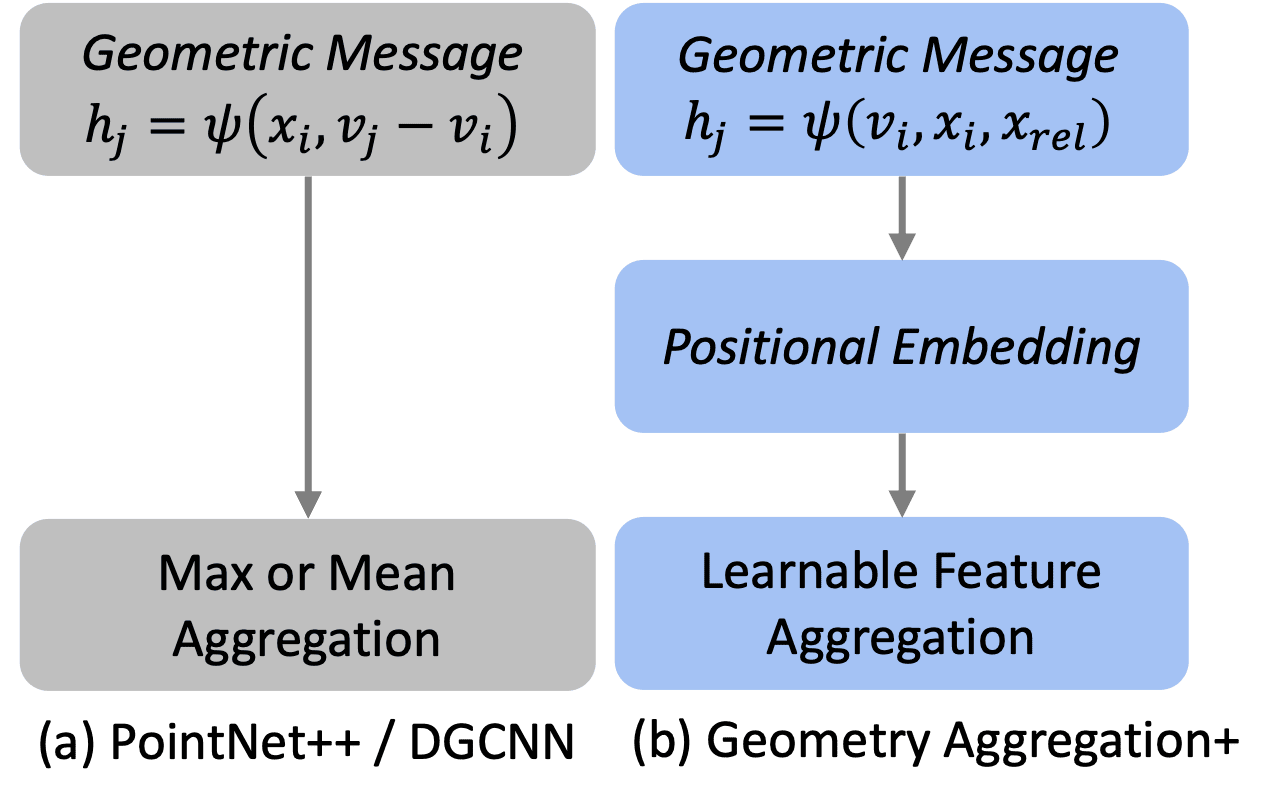}
\caption{Different graph message-passing networks adopted in (a) typical point-based / graph-based learning approaches \citep{qi2017pointnet++,wang2019dynamic} and (b) the geometry aggregation+ (GA+) module.}
\label{fig:gap}
\end{figure}

GA+ achieves this by combining three components, as shown in Fig.~\ref{fig:gap}: (\textit{i}) geometric feature learning through graph message passing, (\textit{ii}) trigonometric positional encoding (PE) to inject relative spatial relationships, and (\textit{iii}) a learnable generalized softmax aggregation that adaptively fuses neighborhood features. This combination enables GA+ to learn both hierarchical and local geometric embeddings when integrated into LMSeg’s dual-pooling framework (Fig.~\ref{fig:architecture}(d)).

\paragraph{Graph message-passing convolution:} 
Following common formulations in point- and graph-based learning \citep{qi2017pointnet++,tailor2021towards,zhao2021point}, GA+ learns a local geometric latent embedding $\mathbf{h}_j$ from the features of a target node $\mathbf{v}_i \in \mathcal{V}$ and its adjacent (1-ring) graph neighbors $\mathbf{v}_j, \mathbf{x}_j \in \mathcal{K}_i$ through following:
\begin{gather}
\label{eq:msg_passg}
    \mathbf{h}_j = \psi(\mathbf{v}_i, \mathbf{x}_i, \mathbf{x}_{rel}), \quad 
    \mathbf{x}_{rel} = \frac{(\mathbf{x}_j - \mathbf{x}_i)}{\sigma + \epsilon},
\end{gather}
 where $\psi$ denotes a shared MLP, $\mathbf{x}_i \in \mathcal{X}$ are the learned mesh features (e.g., mesh normals, RGB texture) of the target node $\mathbf{v}_i$, $\sigma$ is the standard deviation of relative node features, and $\epsilon$ is a small constant ($10^{-5}$) for numerical stability \citep{ma2022rethinking}.

\paragraph{Positional embedding:} 
The barycentric dual graph encodes both topology (via $\mathcal{E}$) and geometry (via $\mathcal{V}$). To better capture fine-grained spatial variation such as sharp edges or curvature discontinuities, we map low-dimensional relative coordinates $\mathbf{v}_{rel} = (\mathbf{v}_j - \mathbf{v}_i)$ into a high-dimensional space using trigonometric PE \citep{tancik2020fourier,zhang2023starting}:
\begin{equation}
\begin{aligned}
\label{eq:pe}
    PE(\mathbf{v}_{rel}, 2c) &= \sin(\alpha \cdot \mathbf{v}_{rel} / \beta^{(2c/d)}), \\
    PE(\mathbf{v}_{rel}, 2c+1) &= \cos(\alpha \cdot \mathbf{v}_{rel} / \beta^{(2c/d)}),
\end{aligned}
\end{equation}
where $\alpha = 100$ controls spatial granularity, $\beta = 1000$ is the base frequency, $c$ is the channel index, and $d$ is the embedding dimension.  
Equation~\ref{eq:pe} is derived by mapping the relative position vector $\mathbf{v}_{rel}$ into a high-dimensional space using sinusoidal basis functions, following the general Fourier feature mapping $\gamma(\mathbf{p}) = \left[ \sin(2^c \pi \mathbf{p}), \cos(2^c \pi \mathbf{p}) \right]_{c=0}^{d/2-1}$. We replace $2^c\pi$ with a continuous scaling term $\alpha / \beta^{(2c/d)}$ to yield a smooth logarithmic progression of frequencies, capturing both coarse and fine geometric variations. The constants $\alpha$ and $\beta$ are selected to balance sensitivity to high-frequency curvature changes with numerical stability for large-scale meshes, following prior work~\citep{tancik2020fourier,zhang2023starting}. The encoded position is fused with the local geometric latent embedding:
\begin{gather}
\label{eq:h_pe}
    \mathbf{h}^{PE}_j = \phi(\mathbf{h}_j \cdot PE(\mathbf{v}_{rel}) + PE(\mathbf{v}_{rel})),
\end{gather}
where $\phi$ is a shared ResMLP. Here, $\mathbf{h}^{PE}_j$ denotes the neighbor feature $\mathbf{h}_j$ enriched with trigonometric positional encoding, enabling the representation to carry both learned feature semantics and explicit high-frequency spatial information.

\paragraph{Learnable generalized softmax aggregation:}
Traditional aggregation operators (e.g., \texttt{mean}, \texttt{max}, \texttt{min}) offer desirable geometric invariance properties but lack adaptability to varying local structures \citep{li2023deepergcn,xu2018how}. To address this inadaptability, GA+ leverages a differentiable generalized softmax aggregation \citep{li2023deepergcn}. Let $s_{ij}=\varphi(\mathbf{h}^{PE}_j)\in\mathbb{R}$ be a learnable scalar score (e.g., a shared linear layer or small MLP applied to $\mathbf{h}^{PE}_j$). Neighborhood weights are then
\begin{equation}
    \omega_{ij} \;=\; \frac{\exp\!\big(t \cdot s_{ij}\big)}{\sum_{n \in \mathcal{K}_i} \exp\!\big(t \cdot s_{in}\big)},
\end{equation}
where $t$ is a learnable inverse temperature controlling the sharpness of attention weights. The aggregation becomes
\begin{equation}
\label{eq:aggr_softmax}
    \mathrm{aggr}_{\mathbf{softmax}}\!\left(\{\mathbf{h}^{PE}_j\}; t\right) \;=\; \sum_{j \in \mathcal{K}_{i}} \omega_{ij}\, \mathbf{h}^{PE}_j.
\end{equation}
This formulation subsumes common aggregation operators as special cases:
\begin{equation}
\omega_{ij} \xrightarrow{\;t\to 0\;} \frac{1}{|\mathcal{K}_i|} \;(\texttt{mean}), 
\quad
\omega_{ij} \xrightarrow{\;t\to +\infty\;} \delta_{j,j^*} \;(\texttt{max}), 
\quad
\omega_{ij} \xrightarrow{\;t\to -\infty\;} \delta_{j,j_*} \;(\texttt{min}),
\end{equation}
where $j^* = \arg\max_{j\in\mathcal{K}_i} s_{ij}$ and $j_* = \arg\min_{j\in\mathcal{K}_i} s_{ij}$.
The temperature $t$ is initialized as a constant and learned via backpropagation, allowing GA+ to interpolate smoothly between global averaging and localized high-saliency aggregation depending on geometric context. In practice, GA+ employs two parallel aggregation operators initialized with $t=0.0$ and $t=1.0$:
\begin{equation}
\label{eq:aggr_softmax_add}
    \mathbf{h}^{aggr}_i \;=\; \mathrm{aggr}_{\mathbf{softmax}}\!\left(\{\mathbf{h}^{PE}_j\};\, t{=}0.0\right) \;+\; \mathrm{aggr}_{\mathbf{softmax}}\!\left(\{\mathbf{h}^{PE}_j\};\, t{=}1.0\right),
\end{equation}
capturing both the global statistical distribution and localized geometric saliency of neighboring features.

\subsection{Hierarchical \& Local Pooling}
\label{sec:pooling}
Efficient pooling is essential for scalable segmentation of large-scale 3D meshes. LMSeg employs a two-stage pooling strategy that balances computational efficiency with geometric fidelity, as shown in Fig.~\ref{fig:architecture}(d):  (\textit{i}) random node sub-sampling to reduce graph size without expensive geometric heuristics, and (\textit{ii}) edge similarity pooling to restore semantically meaningful local neighborhoods.

\paragraph{Random node sub-sampling:} 
At each encoder stage, a uniform random subset $\mathcal{S}$ of nodes from $\mathcal{V}^{l-1}$ is selected to form the coarsened graph:
\begin{equation}
    \mathcal{G}^l(\mathcal{M}) = (\mathcal{V}^l_{\mathcal{S}}, \mathcal{E}^l_{\text{sparse}}), \quad \mathcal{H}^l_{\mathcal{S}} \in \mathbb{R}^{|\mathcal{S}| \times D}.
\end{equation}
Edges are retained only between sampled nodes. This operation has constant-time complexity $\mathcal{O}(1)$ per node, making it significantly more efficient than farthest point sampling (FPS) \citep{qi2017pointnet++} or quadric error metrics (QEM) \citep{garland1997surface}, while maintaining uniform coverage.

\paragraph{Edge similarity pooling:} 
Random sub-sampling can weaken local connectivity and disrupt geodesic relationships. To restore structure, the sparse edge set $\mathcal{E}^l_{\text{sparse}}$ is first augmented with Euclidean nearest neighbors per node, producing $\mathcal{E}^l_{\text{dense}}$. We then prune edges with low cosine similarity between connected node features:
\begin{equation}
\label{eq:cos_similarity}
\mathcal{S}_\mathcal{C}(h_i, h_j) = \frac{h_i \cdot h_j}{\max(\lVert h_i \rVert_2 \cdot \lVert h_j \rVert_2)},
\end{equation}
retaining only edges with similarity above a fixed threshold. This step recovers semantically meaningful neighborhoods with minimal memory overhead.

\paragraph{Hierarchical and local neighborhoods:} 
At each encoder stage, LMSeg constructs two complementary graphs:  
(\textbf{hierarchical}) $\mathcal{E}^l_{\text{hier}}$, linking each downsampled node to its $k$ nearest neighbors in the original full-resolution coordinates, and  
(\textbf{local}) $\mathcal{E}^l_{\text{local}}$, preserving short-range geodesic connections within the pooled graph. HGA+ module operates on $\mathcal{E}^l_{\text{hier}}$ to aggregate long-range, multi-scale context from the original geometry, enabling the network to capture coarse structural patterns across large surface regions. LGA+ module operates on $\mathcal{E}^l_{\text{local}}$ to refine features using high-frequency, fine-scale variations within compact neighborhoods, preserving detail lost during downsampling.  

The outputs of HGA+ and LGA+ are concatenated and updated via a residual MLP:
\begin{equation}
    \mathbf{h}^l = \text{ResMLP}\left( \left[ \mathbf{h}^l_{\text{HGA+}} \ \Vert \ \mathbf{h}^l_{\text{LGA+}} \right] \right),
\end{equation}
where $\Vert$ denotes feature concatenation, and $\mathbf{h}^l_{\text{HGA+}}$, $\mathbf{h}^l_{\text{LGA+}}$ are the respective outputs.  
This dual-module pooling and aggregation ensures explicit multi-scale feature fusion within each encoder stage, with HGA+ contributing global geometric context and LGA+ providing local geometric detail—critical for accurate segmentation of small, irregular, and topologically complex features.

\subsection{Theoretical basis of GA+:}  
GA+ integrates three complementary components—graph message passing, trigonometric positional encoding, and learnable generalized softmax aggregation—to enhance feature learning on the barycentric dual graph $\mathcal{G}(\mathcal{M})$. Graph message passing (Eq.~\ref{eq:msg_passg}) captures local geometric relationships by propagating and transforming features across face-adjacency neighborhoods. Positional encoding (Eq.~\ref{eq:pe}) enriches these interactions with high-frequency spatial information, enabling sensitivity to fine-scale variations such as sharp edges, curvature discontinuities, and localized topological changes. The learnable generalized softmax aggregation (Eq.~\ref{eq:aggr_softmax}) adaptively weights neighborhood features during message passing, providing a continuous interpolation between mean, max, and min operators based on the learned geometric context. Together, these mechanisms increase the expressivity of the learned embeddings, allowing GA+ to capture both coarse structural patterns and subtle high-frequency cues. Within the dual-pooling framework, hierarchical GA+ module (HGA+) leverages this capability to model long-range dependencies, while local GA+ module (LGA+) preserves fine-grained detail. This combination is particularly effective for segmenting small objects in complex urban and natural landscapes, where discriminative features are often sparse, spatially localized, and embedded within geometrically diverse surroundings.

\section{Experiments}
\label{experiments}
We evaluate LMSeg on three large-scale landscape surface mesh datasets: SUM \citep{gao2021sum}, H3D \citep{kolle2021hessigheim} and Budj Bim Wall (BBW) dataset, covering distinct spatial domains to assess both segmentation accuracy and cross-domain robustness. Results are compared against existing learning-based segmentation models, with quantitative and qualitative outcomes reported in Sec.~\ref{sub_sec:results}.

\subsection{Dataset and Pre-processing}
\label{sub_sec:dataset}

\begin{figure*}[!tbp]
\centering
\includegraphics[width=1\textwidth]{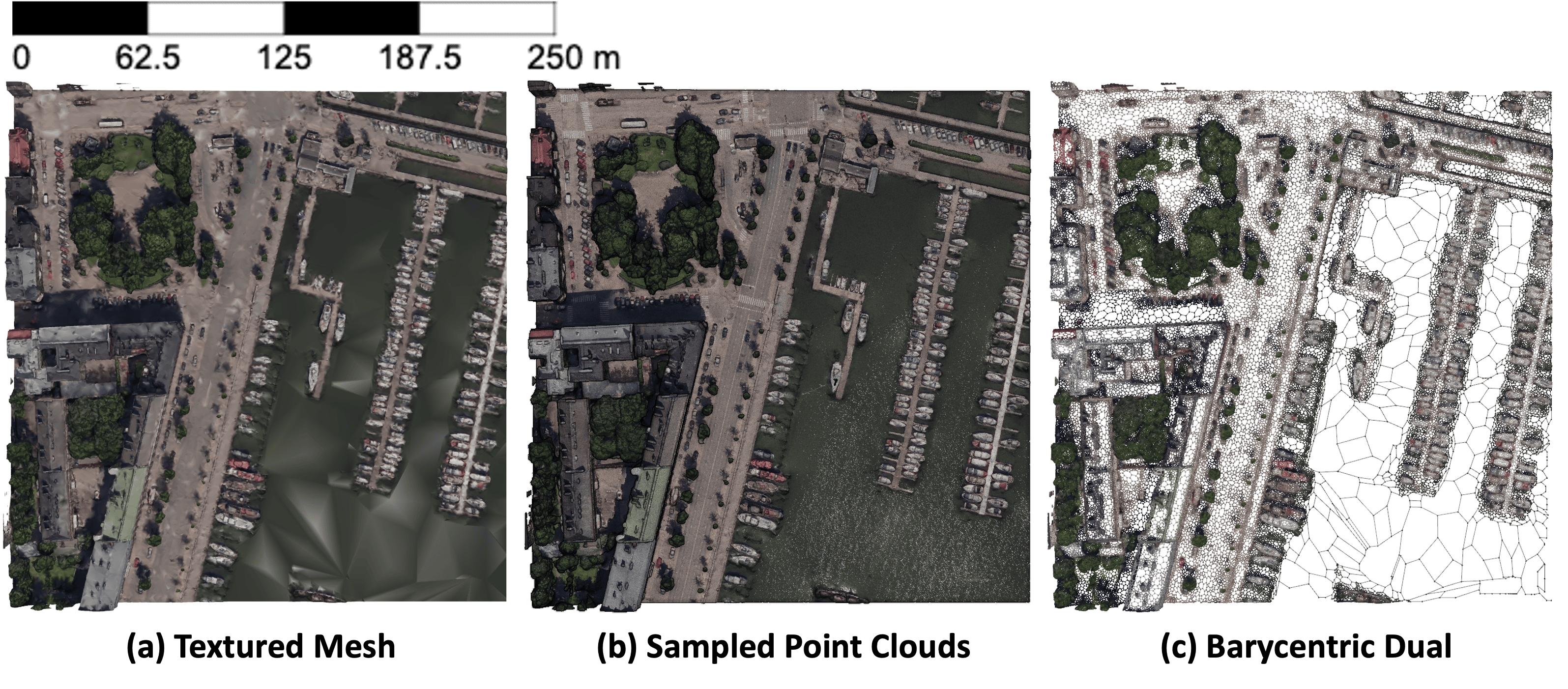}
\caption{(a) Non-uniformly textured mesh, (b) 3D point clouds, and (c) barycentric dual graph of the SUM dataset. The point clouds are densely sampled from the textured mesh at 30 points/m$^2$ for point-based learning models \citep{gao2021sum}.}
\label{fig:sum_data}
\end{figure*}

\textbf{SUM} dataset \citep{gao2021sum} is a public benchmark dataset of semantic urban meshes (Fig.~\ref{fig:sum_data}), covering approximately 8\,km$^2$ of total captured surface area (4\,km$^2$ map area, plus objects) in central Helsinki, Finland. It contains six semantic object classes and an unclassified complement class. The dataset consists of 64 tiles: 40 training, 12 validation, and 12 test tiles, following the official split to ensure comparability with prior works \citep{gao2021sum,gao2023pssnet}. Each tile covers a 250\,m$^2$ map area with a mesh face density of about 6.5 faces/m$^2$.

\begin{figure*}[!tbp]
\centering
\includegraphics[width=1\textwidth]{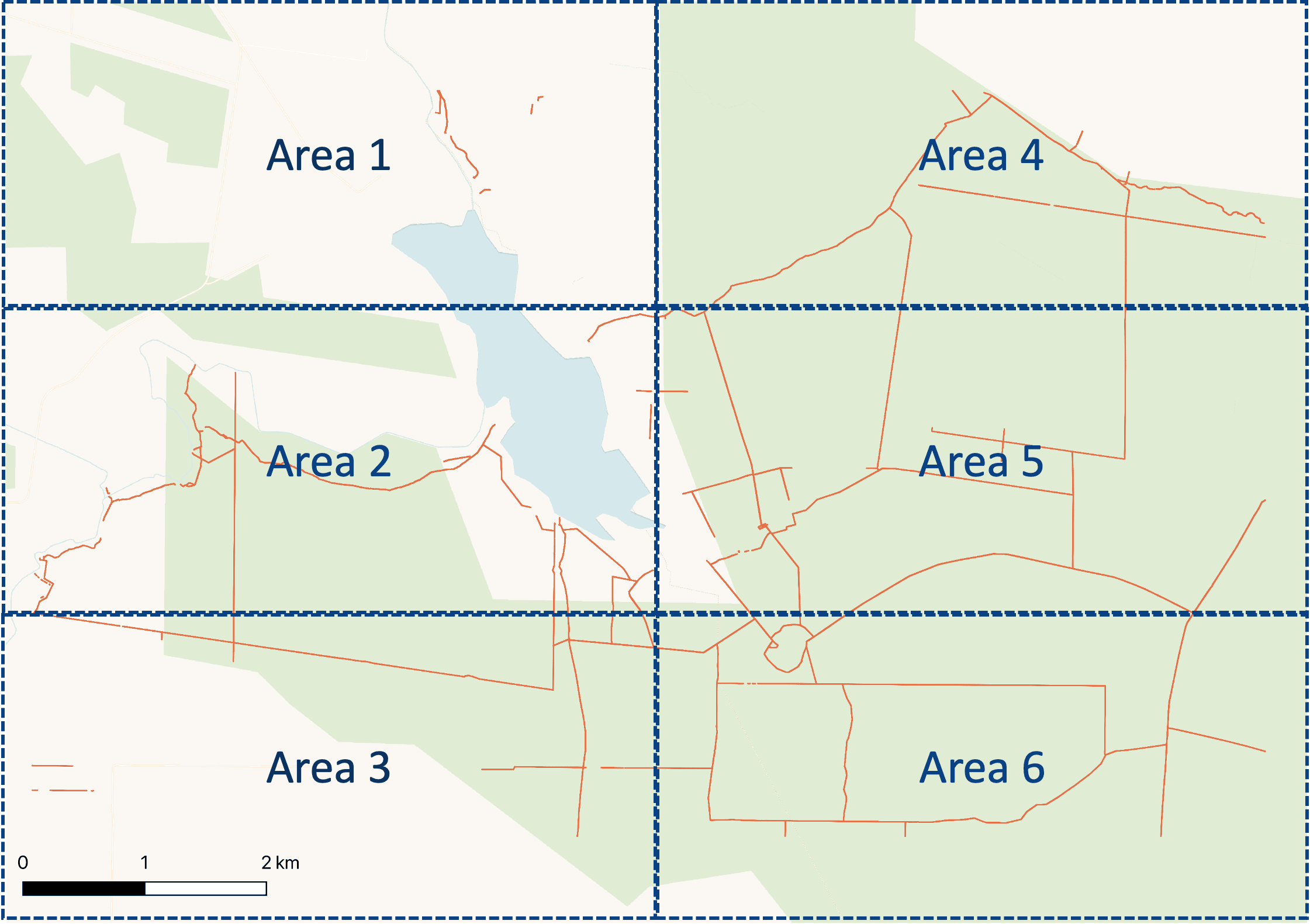}
\caption{Spatial partitioning of the \textit{BudjBimWall} dataset. Orange lines indicate annotated European historic dry-stone walls near Tae Rak (Lake Condah), Victoria, Australia. The number of data samples per area is as follows: Area 1 - 107, Area 2 - 647, Area 3 - 625, Area 4 - 716, Area 5 - 893, and Area 6 - 1008.}
\label{fig:bbw_map}
\end{figure*}

\begin{figure*}[!tbp]
\centering
\includegraphics[width=1\textwidth]{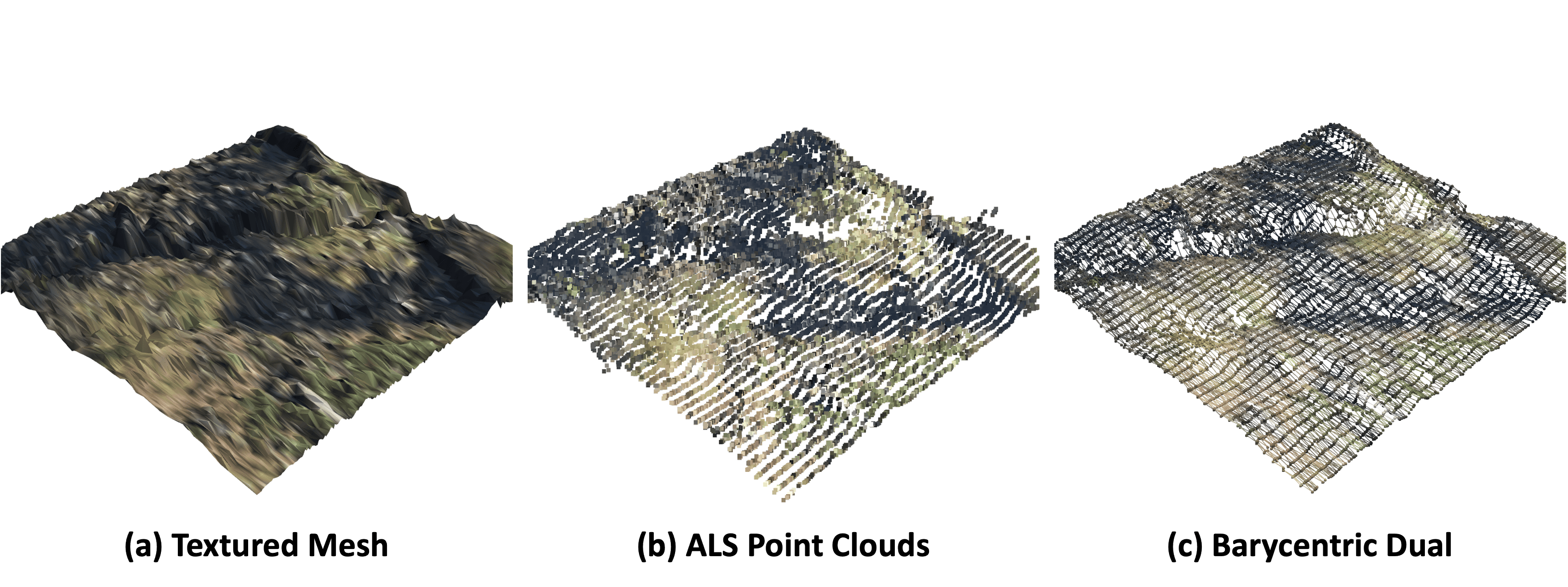}
\caption{(a) Near-uniform textured triangular mesh, (b) 3D lidar point cloud (ALS), and (c) barycentric dual graph of the BBW dataset.}
\label{fig:bbw_dataset}
\end{figure*}

\textbf{BBW} dataset is a lidar-scanned point clouds dataset of the UNESCO World Heritage cultural landscape \citep{smith2019indigenous,bell2019new} in Budj Bim National Park, southwest Victoria, Australia. This region contains one of the highest densities of European historic dry-stone walls in Australia. The dataset, collected in 2020 by the Department of Environment, Land, Water and Planning (Victoria) for the Gunditj Mirring Traditional Owners Corporation, captures the northern portion of the survey area. It is divided into six equal rectangular areas (Fig.~\ref{fig:bbw_map}). Each tile covers 400\,m$^2$ at a face density of $\sim$45 faces/m$^2$ and is semi-manually annotated into binary semantic classes (wall vs. other terrain). 

\begin{figure*}[!tbp]
\centering
\includegraphics[width=1\textwidth]{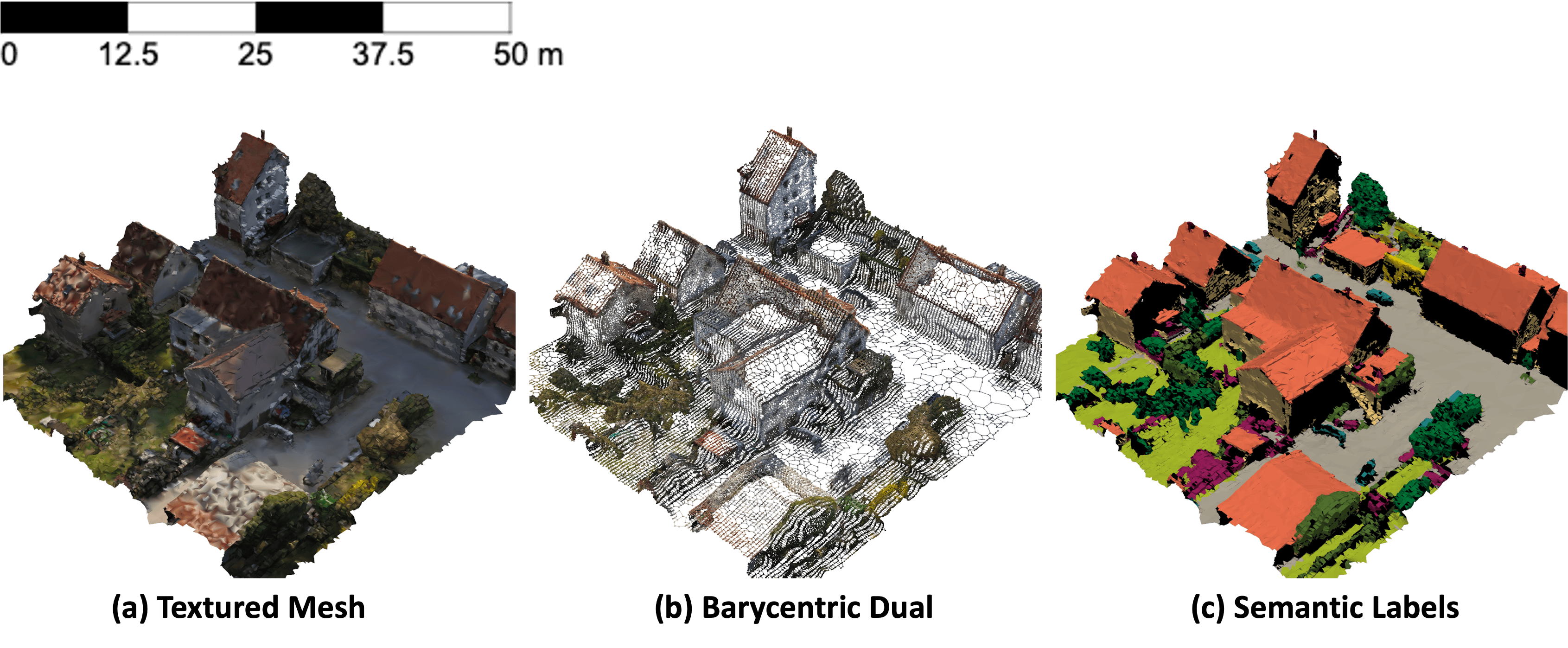}
\caption{(a) Non-uniformly textured mesh, (b) barycentric dual graph and (c) semantic labels of the H3D dataset \citep{kolle2021hessigheim}.}
\label{fig:h3d_data}
\end{figure*}

\textbf{H3D} dataset \citep{kolle2021hessigheim} (Fig.~\ref{fig:h3d_data}) covers approximately 0.19\,km$^2$ of the German village of Hessigheim. It consists of both UAV laser scanning point clouds and textured meshes, annotated into 11 fine-grained semantic classes plus one unlabeled class. Compared to SUM, H3D poses unique challenges: only about 60\% of the mesh surface is labeled, while the remaining 40\% is unlabeled. Moreover, many mesh faces carry incorrect labels due to limitations in transferring annotations from points to mesh triangles, where correspondences may be incomplete or ambiguous. These factors make H3D less suited as a standalone training dataset, as deep learning methods typically require large volumes of consistent annotations. Instead, H3D serves as a challenging benchmark for testing model robustness under conditions of incomplete and noisy supervision.

\textbf{Pre-processing.}  
For SUM and H3D datasets, we use the official textured triangular meshes and apply mesh cleaning before converting them to barycentric dual graphs, removing small defects such as degenerate faces and non-manifold edges. For BBW dataset, we generate textured meshes (Fig.~\ref{fig:bbw_dataset}) from raw lidar tiles using a reproducible pipeline. Terrain point clouds are clipped to area boundaries, subdivided into processing grids, and filtered with a cloth simulation filter (CSF; resolution 0.05\,m, rigidness 1, slope smoothing enabled) \citep{zhang2016easy} to isolate ground points. Ground surfaces are then triangulated in 2D, elevations restored, and meshes simplified to reduce redundancy. Subsequent cleaning removes non-manifold edges, degenerate faces, and unreferenced vertices; merges vertices within tolerance; fills small holes; and reorients normals. These steps explicitly address mesh noise and missing-face artefacts, ensuring geometric integrity for downstream segmentation. Finally, RGB texture and binary wall/terrain masks are mapped to mesh faces by sampling centroids from orthophotos and raster masks, producing enriched meshes with per-face color and labels.

Each mesh $\mathcal{M}$ is converted into its barycentric dual graph $\mathcal{G}(\mathcal{M})$, where nodes represent face barycenters. Node features include normalized 3D coordinates, per-face RGB values, and unit face normals.

\textbf{Data augmentation.}  
To improve robustness and domain generalization, we apply consistent augmentation to SUM, H3D and BBW datasets. Geometric augmentations follow practices from \citet{qian2022pointnext,yang2023surface}: random $z$-axis rotations in $[-180^\circ, 180^\circ]$, small $x$- and $y$-axis rotations in $[-1^\circ, 1^\circ]$, and per-node jitter in $[-0.005, 0.005]$. Photometric augmentations use a lightweight \textit{ColorJitter} module that randomly adjusts brightness, contrast, saturation, and hue, applies probabilistic grayscale or zero-color dropout, and adds Gaussian noise. These simulate lighting variation, sensor noise, and partial texture loss, enhancing generalization to new spatial domains.

\textbf{Splitting strategy.}  
For the SUM dataset, we retain the official 40/12/12 split for comparability. For the H3D dataset, several temporal versions of the mesh are available; we use the March 2018 epoch for model training and validation, which is also adopted in the original H3D benchmark. This choice ensures consistency with prior benchmark studies \citep{thomas2019kpconv,qi2017pointnet++,landrieu2018large,gao2023pssnet} and facilitates fair comparison. For the BBW dataset, we adopt a six-fold \textit{leave-one-area-out} cross-validation, testing on one held-out geographic area per fold while training and validating (80/20 split) on the remaining five. This design covers diverse terrain types and provides a statistically robust estimate of model performance on unseen spatial regions. Evaluating LMSeg on these spatially and visually distinct datasets—urban aerial meshes (SUM), high-resolution UAV-based textured meshes (H3D), and lidar-derived rural heritage landscapes (BBW)—enables a more direct assessment of robustness and transferability across domains.

\subsection{Implementation}
We train LMSeg using AdamW \citep{loshchilov2017decoupled} with a weight decay of $1\!\times\!10^{-4}$. The initial learning rate is set to 0.001 for SUM (batch size 1) and BBW (batch size 8), and 0.005 for H3D (batch size 1), following 3D segmentation training setups in \citet{qian2022pointnext}. Batch sizes reflect GPU memory constraints. A cosine annealing scheduler is used to adaptively decay the learning rate during training. For loss functions, we adopt multiclass cross-entropy with label smoothing \citep{muller2019does} for SUM and H3D and binary cross-entropy with label smoothing for BBW. To mitigate class imbalance, we compute normalized class-weight vectors from dataset label distributions and incorporate them into the loss functions, following \citet{yang2023surface}. 

For SUM, we report mean Intersection-over-Union (mIoU) and mean class accuracy (mAcc). For H3D, we follow the benchmark protocol and report mean F1-score and overall accuracy (O.A.). For BBW, we evaluate using mIoU and F1-score, reflecting its binary classification setting. All models are implemented in PyTorch Geometric \citep{fey2019fast} and trained on a Linux server equipped with an NVIDIA A100-80Q GPU.

\subsection{Results}
\label{sub_sec:results}

\begin{table*}[!tbp]
\tiny
\centering
\caption{3D semantic segmentation results on SUM dataset. Proportions of surface area (per semantic object class) of test set are displayed next to each category in (\%). Model results are collected from the original papers reporting the performance on public benchmarks SUM \citep{gao2021sum}, TexturalSG \citep{yang2023surface}, PSSNet \citep{gao2023pssnet} and UMeshSeg \citep{liu2024umeshsegnet}. We compare the performance of LMSeg (average performances over ten runs) with existing learning-based segmentation methods. Results reported are per-class IoU (\%), mean IoU (\%), overall accuracy (\%) and mean accuracy (\%). Model inference time (minute) is measured on the SUM test dataset. Best results marked in bold. $\dagger$ indicates a single-run result reported by the original paper; variance was not available. $\ast$ indicates models tested on a Linux computing server with a 32-core Intel Xeon Platinum CPU.}
\begin{tabular}{clccccccccccrr}
\toprule
Modality & Method & \shortstack{Terra.\\ (23.6\%)} & \shortstack{H-veg.\\ (14.3\%)} & \shortstack{Build. \\ (50.7\%)} & \shortstack{Water \\ (4.8\%)} & \shortstack{Vehic. \\ (1.5\%)} & \shortstack{Boat \\ (2.1\%)} & mIoU & O.A. & mAcc. & Sampling & \shortstack{Params. \\ (M)} & \shortstack{$t$ inf.\\ (Min)} \\
\midrule
\multirow{5}{*}{\rotatebox[origin=c]{90}{Point Clouds}} 
& PointNet \citep{qi2017pointnet} & 56.3 & 14.9 & 66.7 & 83.8 & 0.0 & 0.0 & 36.9 $\pm$ 2.3& 71.4 $\pm$ 2.1 & 46.1 $\pm$ 2.6 & - & 3.6 & \textbf{1} \\
& RandLANet \citep{hu2020randla} & 38.9 & 59.6 & 81.5 & 27.7 & 22.0 & 2.1 & 38.6 $\pm$ 4.6 & 74.9 $\pm$ 3.2 & 53.3 $\pm$ 5.1 & Random & 1.3 & 52 \\
& SPG \citep{landrieu2018large} & 56.4 & 61.8 & 87.4 & 36.5 & 34.4 & 6.2 & 47.1 $\pm$ 2.4 & 79.0 $\pm$ 2.8 & 64.8 $\pm$ 1.2 & - & - & 26 \\
& PointNet++ \citep{qi2017pointnet++} & 68.0 & 73.1 & 84.2 & 69.9 & 0.5 & 1.7 & 49.5 $\pm$ 2.1 & 85.5 $\pm$ 0.9 & 57.8 $\pm$ 1.8 & FPS & \textbf{1.0} & 3 \\
& KPConv \citep{thomas2019kpconv} & 86.5 & 88.4 & 92.7 & 77.7 & 54.3 & 13.3 & 68.8 $\pm$ 5.7 & 93.3 $\pm$ 1.5 & 73.7 $\pm$ 5.4 & FPS & 15.0 & 42\\
\midrule
\multirow{5}{*}{\rotatebox[origin=c]{90}{Mesh}} 
& RF-MRF \citep{rouhani2017semantic} & 77.4 & 87.5 & 91.3 & 83.7 & 23.8 & 1.7 & 60.9 $\pm$ 0.0 & 91.2 $\pm$ 0.0 & 65.9 $\pm$ 0.0 & - & - & 15 \\
& SUM-RF \citep{gao2021sum} & 83.3 & 90.5 & 92.5 & 86.0 & 37.3 & 7.4 & 66.2 $\pm$ 0.0 & 93.0 $\pm$ 0.0 & 70.6 $\pm$ 0.0 & - & - & 18 \\
& TexturalSG \citep{yang2023surface} & \textbf{88.2} & 91.0 & 92.9 & \textbf{90.1} & 47.6 & 18.9 & 71.5 $\pm$ 1.9 & 94.1 $\pm$ 1.2 & - & Random & - & - \\
& PSSNet \citep{gao2023pssnet} & 84.9 & 90.6 & \textbf{93.9} & 84.3 & 50.9 & \textbf{32.3} & 72.8 $\pm$ 2.0 & 93.8 $\pm$ 0.4 & 79.2 $\pm 3.0 $ & - & - & 62 \\
& UMeshSeg \citep{liu2024umeshsegnet}$^{\dagger}$  & 67.3 & 84.2 & 86.7 & 50.7 & 34.2 & 20.1 & 57.3 $\pm$ 0.0 & 87.6 $\pm$ 0.0 & 69.5 $\pm$ 0.0 & - & - & - \\
\cmidrule(l){2-14}
\rowcolor{lightgray}
\cellcolor{white} & LMSeg (Ours-avg) & 83.5 & \textbf{94.3} & 93.6 & 72.8 & \textbf{75.8} & 30.0 & \textbf{75.1} $\pm 0.3 $ & \textbf{94.5} $\pm 0.07$ & \textbf{81.8} $\pm 0.4 $ & Random & 2.4 & $2^{\ast}$ \\
\bottomrule
\end{tabular}
\label{tab:sum_result}
\end{table*}

\begin{figure*}[!tbp]
\centering
\includegraphics[width=1\textwidth]{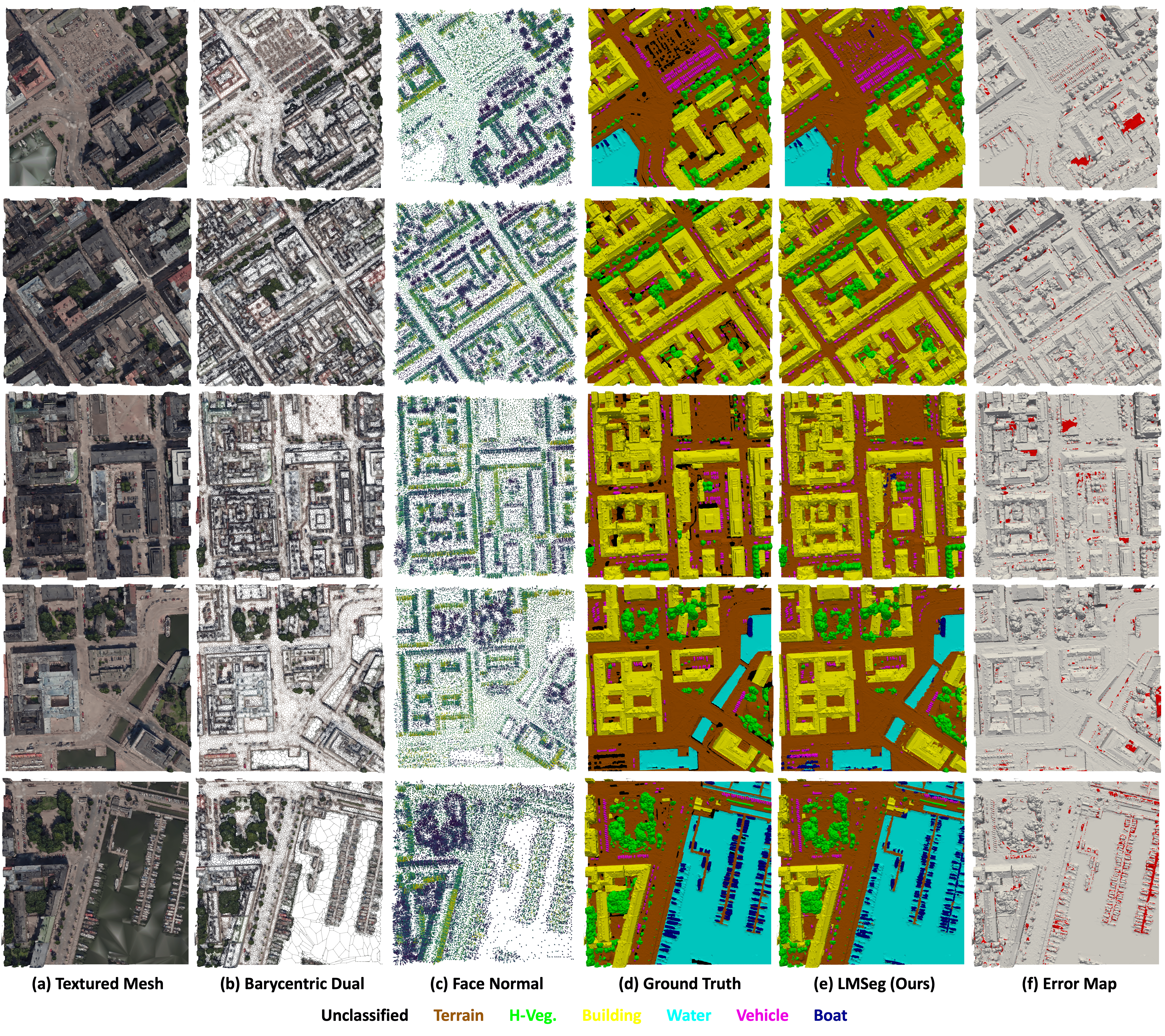}
\caption{Qualitative performance of LMSeg on SUM dataset. Regions without ground-truth labels are displayed in black.}
\label{fig:sum_preds}
\end{figure*}

\paragraph{Performance on SUM:}  
Table~\ref{tab:sum_result} reports LMSeg’s performance on the SUM benchmark, where it achieves 75.1\% mIoU and 81.8\% mAcc averaged over ten runs.  

Compared with early point-based methods \citep{qi2017pointnet,qi2017pointnet++,hu2020randla,thomas2019kpconv} trained on 3D point clouds sampled from textured mesh faces (Fig.~\ref{fig:sum_data}(b)), LMSeg achieves consistent improvements in both mIoU and mAcc. Point-based methods process irregular point sets without explicitly leveraging mesh connectivity, which can result in loss of surface continuity and neighborhood consistency during point sampling. In contrast, LMSeg directly operates on the barycentric dual graph of the mesh, preserving topological relationships and geometric context throughout message passing.  

It also surpasses traditional mesh-based approaches such as RF-MRF \citep{rouhani2017semantic} and SUM-RF \citep{gao2021sum}, which rely on over-segmented superfacets combined with hand-crafted geometric and textural descriptors followed by probabilistic inference. These methods are not end-to-end trainable and require manual feature engineering, whereas LMSeg learns feature representations directly from raw mesh attributes via differentiable GA+ layers, enabling unified feature learning and classification in a single framework.  

More recent deep mesh segmentation networks further advance the state of the art. TexturalSG \citep{yang2023surface} incorporates surface-graph modeling of textured meshes, but rely heavily on high-quality texture maps and color information. LMSeg achieves a slightly higher overall accuracy (94.5\% vs.\ 94.1\%) and substantially higher IoU on small-object categories (Vehic.: 75.8\% vs.\ 47.6\%, Boat: 30.0\% vs.\ 18.9\%), reflecting its ability to capture localized, high-frequency geometric features even in the absence of texture. PSSNet \citep{gao2023pssnet} adopts a two-stage pipeline based on superfacet partitioning and edge-dependency modeling, achieving a comparable mIoU (72.8\% vs.\ 75.1\%), but LMSeg avoids the complexity of multi-stage processing while delivering more balanced class-wise performance and lower variance, indicating greater stability. Against the most recent UMeshSegNet \citep{liu2024umeshsegnet}, which employs multi-scale mesh convolution but is computationally heavier, LMSeg achieves a substantial gain in mIoU (75.1\% vs.\ 57.3\%) and consistently higher IoU across all small-object classes, underscoring the efficiency and discriminative power of its geometry-aware message passing and adaptive feature aggregation. 

Across semantic classes, LMSeg attains competitive IoU in four of six categories—H-veg. (94.3\%), Build. (93.6\%), Vehic. (75.8\%), and Boat (30.0\%)—demonstrating effectiveness on both large-scale and small-scale classes. The disparity between the Vehicle and Boat categories can be attributed to differences in graph node density and feature separability: in SUM, boats occupy a smaller surface area, yielding fewer mesh faces and sparser dual-graph neighborhoods, while also exhibiting high feature ambiguity due to similar surface normals and color distributions with surrounding water. In contrast, vehicles appear more frequently on land surfaces with stronger geometric and contextual separation from their background, allowing GA+’s geometry-aware message passing to extract more distinctive embeddings. These observations suggest that while LMSeg’s hierarchical–local dual-pooling design is effective for many small objects, extreme sparsity and high background similarity remain challenging, thus motivating future work on class-balanced sampling and feature augmentation for underrepresented mesh categories.  

Qualitative examples in Fig.~\ref{fig:sum_preds} confirm LMSeg’s capability to segment both large structures and fine-scale objects, with most residual errors occurring at ambiguous boundaries (e.g., water–terrain transitions) where annotation inconsistencies are known to occur \citep{gao2021sum}. We also note that the SUM dataset contains incomplete annotations for some small-object instances, which may lead to under- or over-estimation of per-class IoU. This limitation means that LMSeg’s improvements in small-object segmentation are likely underrepresented in the reported metrics. To mitigate this bias, we present complementary evidence in controlled ablations (Sec.~\ref{abla_hgap_lgap}) showing consistent gains from HGA+ and LGA+ modules in capturing fine-grained structures

\begin{table*}[!tbp]
\centering
\tiny
\caption{3D semantic segmentation results on the BBW dataset. The dataset contains two semantic classes: Wall (11.4\% of the surface area) and Terrain (88.5\%). Performance is reported as mean Intersection-over-Union (mIoU, \%) and F1-score (\%). All models, including baselines, were evaluated on the same Linux computing server with a 32-core Intel Xeon Platinum CPU to ensure a fair comparison. Results are averaged over 6-fold cross-validation across six areas (Area 1–Area 6). LMSeg achieves the best overall performance with 76.8\% mean F1 and 62.4\% mean IoU. Best results are shown in bold.}
\resizebox{\textwidth}{!}{
\begin{tabular}{l *{6}{cc} cc}
\toprule
\multirow{2}{*}{Model} 
& \multicolumn{2}{c}{Area1} & \multicolumn{2}{c}{Area2} & \multicolumn{2}{c}{Area3}
& \multicolumn{2}{c}{Area4} & \multicolumn{2}{c}{Area5} & \multicolumn{2}{c}{Area6}
& \multirow{2}{*}{Sampling} & \multirow{2}{*}{\makecell{Params. \\ (M)}} \\
\cmidrule(lr){2-3}\cmidrule(lr){4-5}\cmidrule(lr){6-7}\cmidrule(lr){8-9}\cmidrule(lr){10-11}\cmidrule(lr){12-13}
& F1 & mIoU & F1 & mIoU & F1 & mIoU & F1 & mIoU & F1 & mIoU & F1 & mIoU &  &  \\
\midrule
PointNet & 38.5 & 23.9 & 46.5 & 30.3 & 63.1 & 46.1 & 44.6 & 28.7 & 39.5 & 24.6 & 46.6 & 30.4 & -- & 3.6 \\
RandLANet & 71.0 & 55.0 & 73.4 & 57.9 & 81.2 & 68.4 & 70.5 & 54.4 & 70.0 & 53.8 & 75.3 & 60.3 & Random & 1.3 \\
PointNet++ & 66.3 & 49.6 & 67.6 & 51.0 & 77.9 & 63.8 & 65.8 & 49.1 & 64.6 & 47.7 & 70.4 & 54.4 & FPS & \textbf{1.0} \\
KPConv & 67.7 & 51.1 & 69.1 & 52.8 & 77.3 & 63.0 & 65.9 & 49.2 & 62.3 & 45.2 & 67.2 & 50.6 & FPS & 15.0 \\
PointTransformer & 67.5 & 50.9 & 69.4 & 53.1 & 79.0 & 65.3 & 68.1 & 51.6 & 68.1 & 51.6 & 72.6 & 57.0 & FPS & 4.9 \\
DeeperGCNs & 50.4 & 33.7 & 50.2 & 33.5 & 70.0 & 53.9 & 46.4 & 30.2 & 38.2 & 23.6 & 55.7 & 38.6 & -- &  1.9\\
Graph U-Net & 38.2 & 23.6 & 52.3 & 35.5 & 68.8 & 52.4 & 42.3 & 26.8 & 40.0 & 25.0 & 54.7 & 37.7 & Top$k$ & 0.85 \\
\rowcolor{lightgray}
LMSeg (Ours) & \textbf{75.8} & \textbf{61.0} & \textbf{75.6} & \textbf{60.8} & \textbf{82.7} & \textbf{70.5} & \textbf{75.0} & \textbf{60.0} & \textbf{73.6} & \textbf{58.2} & \textbf{78.2} & \textbf{64.2} & Random & 2.4 \\
\bottomrule
\end{tabular}
}
\label{tab:bbw_result}
\end{table*}

\begin{figure*}[!tbp]
\centering
\includegraphics[width=1\textwidth]{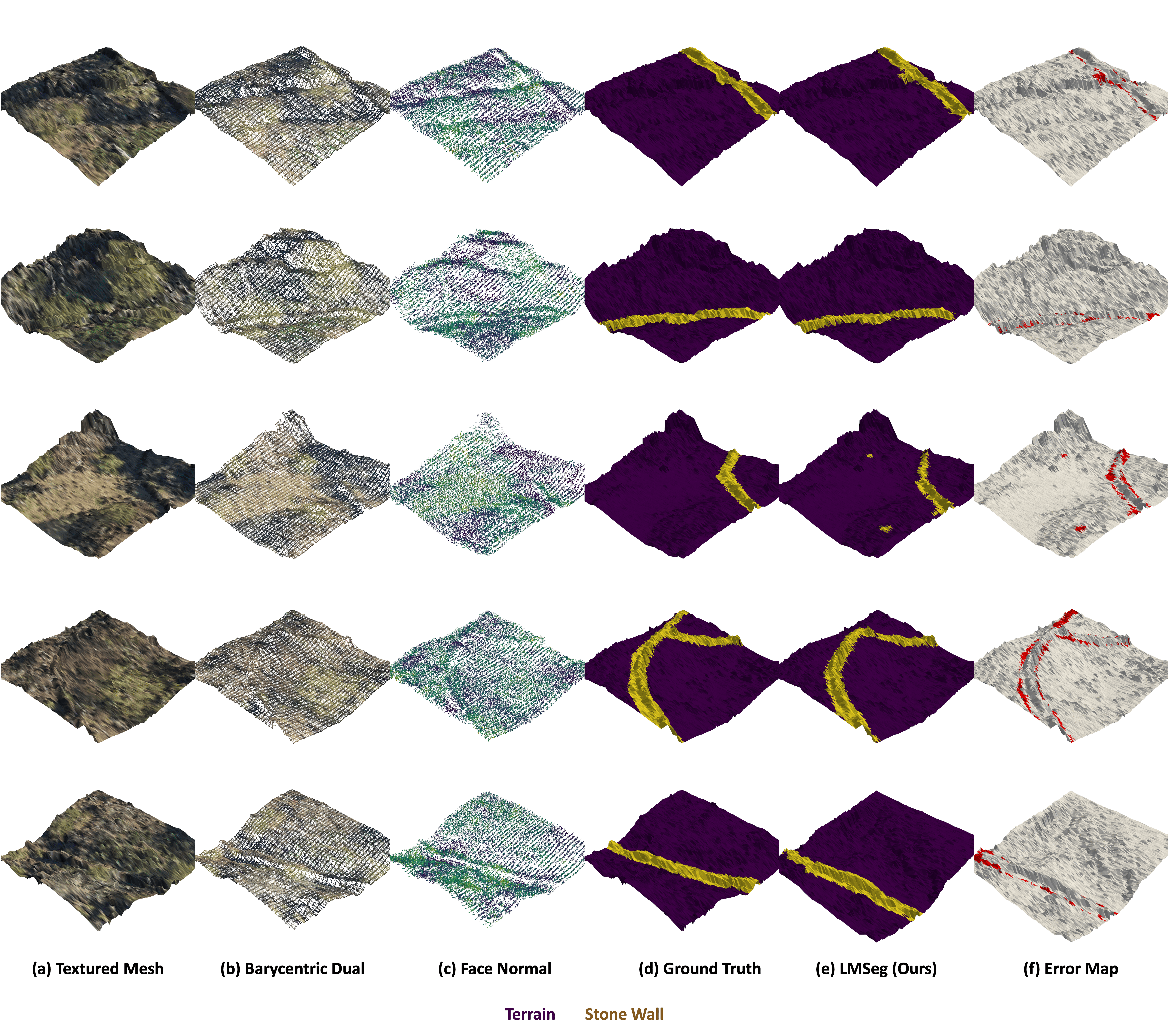}
\caption{Qualitative performance of LMSeg on BBW dataset.}
\label{fig:bbw_preds}
\end{figure*}

\paragraph{Performance on BBW:}  
Table~\ref{tab:bbw_result} summarizes results on the BBW dataset, which contains only two semantic classes (Wall and Terrain) with a highly imbalanced surface area distribution (11.4\% vs.\ 88.5\%). LMSeg achieves the best overall performance, with 76.8\% mean F1 and 62.4\% mean IoU across six areas, consistently outperforming both point-based and graph-based baselines.

Earlier point-based methods such as PointNet \citep{qi2017pointnet}, PointNet++ \citep{qi2017pointnet++}, and RandLANet \citep{hu2020randla} rely on raw point cloud sampling, where FPS or random strategies may discard critical boundary points. For instance, PointNet++ reaches only 49.6--63.8\% mIoU across different areas, while RandLANet performs better (55.0--68.4\% mIoU) but still underperforms LMSeg in all six regions. More advanced point-based kernels like KPConv \citep{thomas2019kpconv} and PointTransformer \citep{zhao2021point} improve local geometric reasoning, with PointTransformer in particular leveraging transformer-based self-attention to model long-range dependencies. However, even these methods remain constrained by redundant resampling of dense terrain points and do not exploit mesh connectivity, remaining consistently below LMSeg’s range of 58.2--70.5\% mIoU.  

Graph-based baselines provide complementary perspectives for comparison. DeeperGCNs \citep{li2023deepergcn} employ learnable aggregation functions and deeper architectures to enhance expressivity, while Graph U-Net \citep{gao2019graph} introduces a hierarchical U-Net–style encoder–decoder design for graph data. Despite their architectural sophistication, both approaches lack explicit geometry-aware encodings tailored to meshes and instead rely on generic graph pooling (e.g., Top-$k$ in Graph U-Net), resulting in weaker boundary preservation (e.g., 23.6--53.9\% mIoU across areas). By contrast, LMSeg directly leverages the mesh structure without resampling or hand-crafted descriptors. Its GA+ module integrates geometry-aware message passing with trigonometric positional encoding and generalized softmax aggregation, while its dual hierarchical–local pooling ensures both global context capture and local boundary refinement. This design allows LMSeg to achieve the best performance across all areas, e.g., 70.5\% mIoU in Area3 and 64.2\% in Area6, outperforming the strongest point-based (65.3\%) and graph-based (53.9\%) competitors by substantial margins.  

As the qualitative examples shown in Fig.~\ref{fig:bbw_preds}, residual errors in BBW mainly occur at wall–terrain boundaries, where extreme class imbalance and local topographic ambiguity make separation difficult. Terrain surfaces dominate the mesh and generate dense neighborhoods, while walls are narrow and discontinuous, leading to sparse node representations and weaker contextual cues. As a result, LMSeg occasionally misclassifies wall fragments adjacent to steep slopes as terrain. These limitations suggest future directions in class-balanced sampling or boundary-sensitive loss functions to better handle underrepresented wall features.

\begin{table}[ht]
\small
\centering
\caption{3D semantic segmentation results on the H3D dataset. The dataset contains eleven semantic classes: C00: \textbf{Low Vegetation}, C01: \textbf{Impervious Surface}, C02: \textbf{Vehicle}, C03: \textbf{Urban Furniture}, C04: \textbf{Roof}, C05: \textbf{Facade}, C06: \textbf{Shrub}, C07: \textbf{Tree}, C08: \textbf{Soil/Gravel}, C09: \textbf{Vertical Surface}, and C10: \textbf{Chimney}. Model performances are collected from the public H3D benchmarks \citep{kolle2021hessigheim}: PointNet++, SPG, KPConv and PSSNet. We compare the performance of these methods on H3D using per-class F1-score (\%), mean F1-score (mF1, \%), and overall accuracy (O.A., \%). Best results are shown in bold.}
\label{tab:h3d_comparison}
\begin{tabular}{lcccccccccccccc}
\toprule
Method & C00 & C01 & C02 & C03 & C04 & C05 & C06 & C07 & C08 & C09 & C10 & mF1 & O.A. & \shortstack{Params. \\ (M)} \\
\midrule
PointNet++ & 57.5 & 70.7 & 0.0 & 2.1 & 37.5 & 36.6 & 0.3 & 43.3 & 0.0 & 3.4 & 0.0 & 22.9 & 46.8 & \textbf{1.0} \\
SPG & 54.4 & 53.0 & 5.6 & 33.1 & 74.9 & 75.3 & 0.1 & 92.4 & 21.0 & 56.1 & 0.0 & 40.5 & 56.5 & -  \\
KPConv & 64.5 & 64.5 & \textbf{66.0} & 47.3 & \textbf{90.0} & 71.3 & \textbf{57.0} & 91.4 & 12.6 & 58.8 & 27.2 & 59.1 & 70.7 & 15.0  \\
PSSNet & \textbf{84.9} & \textbf{86.4} & 49.2 & 39.5 & 88.4 & 72.3 & 41.3 & 89.9 & \textbf{40.9} & \textbf{70.9} & 60.5 & \textbf{65.8} & \textbf{79.3} & -  \\
\rowcolor{lightgray}
LMSeg (Ours) & 74.8 & 73.0 & 37.0 & \textbf{64.1} & 88.5 & \textbf{78.8} & 46.4 & \textbf{92.7} & 0.1 & 50.4 & \textbf{79.6} & 62.3 & 78.4 & 2.4 \\
\bottomrule
\end{tabular}
\label{tab:h3d_result}
\end{table}

\begin{figure*}[!tbp]
\centering
\includegraphics[width=1\textwidth]{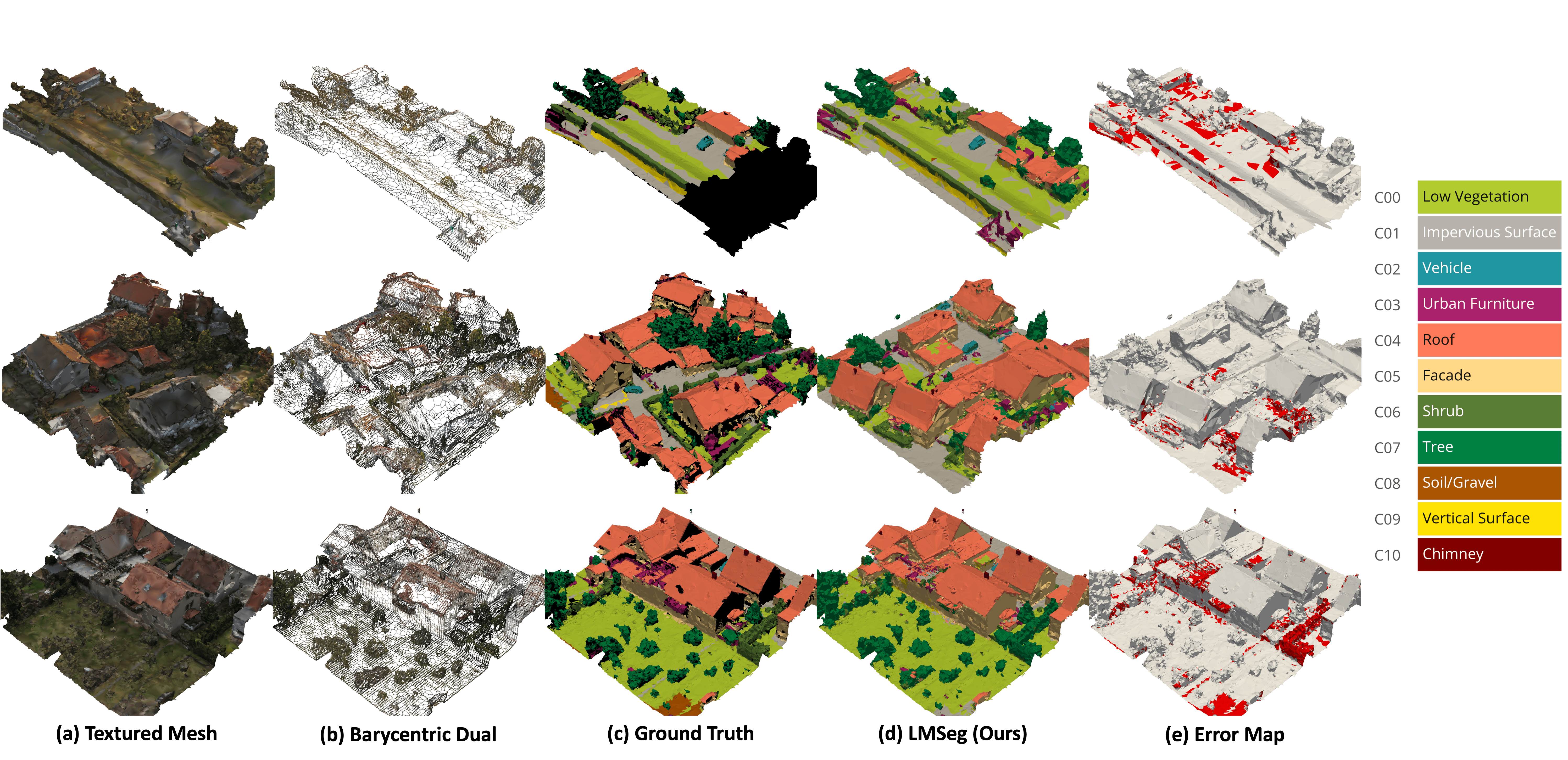}
\caption{Qualitative results of LMSeg on the H3D dataset. Regions without ground-truth labels are displayed in black.}
\label{fig:h3d_preds}
\end{figure*}

\paragraph{Performance on H3D:}
Table~\ref{tab:h3d_result} presents results on the H3D dataset. LMSeg achieves competitive overall performance (62.3\% mF1 and 78.4\% O.A.) with a compact 2.4M parameter model, closely approaching the best-performing PSSNet (65.8\% mF1 and 79.3\% O.A.) while being far lighter than KPConv (15.0M). Class-wise, LMSeg performs particularly well on categories requiring fine-scale structural reasoning: \textbf{Urban Furniture} (C03: 64.1\%), \textbf{Chimney} (C10: 79.6\%), \textbf{Tree} (C07: 92.7\%), and \textbf{Facade} (C05: 78.8\%). This aligns with our findings on the SUM dataset, where LMSeg also performs strongly on small/structured urban objects (e.g., Vehicles), suggesting that the HGA+/LGA+ design effectively balances global context with local boundary refinement across urban scenes.

LMSeg underperforms on \textbf{Soil/Gravel} (C08: 0.1\%), where PSSNet is strongest. PSSNet leverages a two-stage pipeline that over-segments meshes into planar and non-planar patches, aligning object boundaries with planar breaks before classification. This strategy is particularly effective at segmenting \emph{Soil/Gravel} (C08) from visually and geometrically similar classes such as \emph{Low Vegetation} (C00) and \emph{Impervious Surface} (C01). In contrast, LMSeg operates directly at the face level without superfacet pre-grouping; when labels are sparse or noisy, fine-grained ground-class transitions with similar normals/colors can blur, depressing C08 performance. Qualitative results in Fig.~\ref{fig:h3d_preds} reflect this pattern: strong delineation on structured, protruding objects, with occasional confusion among ground-like classes.

\section{Discussion} 
\label{discussion} 
We conduct an ablation study on LMSeg to identify the contributions of its effective components on the SUM and BBW datasets, as summarized in Tab.~\ref{tab:ablated_model}. The H3D dataset is omitted due to noisy labels from imperfect point-to-mesh transfers and its close similarity to SUM in urban semantic attributes, making it unsuitable for reliably assessing module contributions.

\begin{table*}[!tbp]
\small
\centering
\caption{Ablated models tested in the ablation study.} 
\begin{tabular}{lc}
\toprule
Ablated Model & Description \\
\midrule
SUM- and BBW-A & \makecell[l]{Without HGA+ modules (no hierarchical context aggregation).} \\
SUM- and BBW-B & \makecell[l]{Without LGA+ modules, incl. edge similarity pooling (no local refinement).} \\
SUM- and BBW-C & \makecell[l]{With fixed mean/max operators only (no learnable feature aggregation).} \\
\midrule
SUM- and BBW-D & \makecell[l]{Reduced HGA+ neighborhood size to $k=8$ (default $k=32$).} \\
SUM- and BBW-E & \makecell[l]{Reduced HGA+ neighborhood size to $k=16$ (default $k=32$).} \\
\midrule
SUM- and BBW-F & \makecell[l]{Without face and vertex RGB features (geometry only).} \\
SUM- and BBW-G & \makecell[l]{Without face and vertex normals (texture only).} \\
\midrule
SUM- and BBW-H & \makecell[l]{Reduced initial feature dimension to $D=36$ (default $D=72$).} \\
SUM- and BBW-I & \makecell[l]{Learning rate increased to $lr=0.005$ (default $lr=0.001$).} \\
SUM- and BBW-J & \makecell[l]{Learning rate decreased to $lr=0.0005$ (default $lr=0.001$).} \\
\bottomrule
\end{tabular}
\label{tab:ablated_model}
\end{table*}

\begin{table*}[!tbp]
\small
\centering
\caption{Ablation study of LMSeg on SUM dataset. $\Delta$ refers to the performance (mIoU) difference of ablated models compared to LMSeg.}
\begin{tabular}{lcccccccccc}
\toprule
Model & $k$-nbrs. & Encoder & Terra. & H-veg. & Build. & Water & Vehic. & Boat & mIoU & $\Delta$ \\
\midrule
SUM-A & 3$^\ast$ & LGA+ & 75.7 & 92.3 & 91.2 & 21.8 & 58.9 & 11.2 & 58.5 & -16.6 \\
SUM-B & 32 & HGA+ & 81.8 & 94.1 & 93.3 & 67.8 & 74.8 & 27.8 & 73.2 & -1.8 \\
SUM-C & 32 (3$^\ast$) & \makecell{HGA \& LGA} & 79.0 & 93.8 & 92.8 & 54.3 & 69.5 & 24.6 & 69.0 & -6.1 \\
\midrule
SUM-D & 8 (3$^\ast$) & \makecell{HGA+ \& LGA+} & 81.1 & 93.8 & 92.5 & 68.8 & 71.5 & 25.6 & 72.2 & -2.9 \\
SUM-E & 16 (3$^\ast$) & \makecell{HGA+ \& LGA+} & 82.2 & 94.1 & 93.1 & 70.5 & 73.2 & 27.5 & 73.5 & -1.6 \\
\midrule
SUM-F & 32 (3$^\ast$) & \makecell{HGA+ \& LGA+} & 80.9 & 92.1 & 92.8 & 45.6 & 69.8 & 24.0 & 67.5 & -7.6 \\
SUM-G & 32 (3$^\ast$) & \makecell{HGA+ \& LGA+} & 81.7 & 94.0 & 92.9 & 71.9 & 73.7 & 28.9 & 73.9 & -1.2 \\
\midrule
SUM-H & 32 (3$^\ast$) & \makecell{HGA+ \& LGA+} & 80.4 & 93.2 & 92.3 & 63.9 & 73.1 & 26.0 & 71.5 & -3.6 \\
SUM-I & 32 (3$^\ast$) & \makecell{HGA+ \& LGA+} & 83.1 & 94.4 & 93.0 & 71.6 & 75.2 & 28.2 & 74.2 & -0.9 \\
SUM-J & 32 (3$^\ast$) & \makecell{HGA+ \& LGA+} & 82.7 & 94.0 & 93.4 & 70.1 & 74.5 & 30.7 & 74.2 & -0.9 \\
\midrule
\rowcolor{lightgray}
\shortstack{\makecell[l]{LMSeg \\ (Ours-avg)}} & 32 (3$^\ast$) & \makecell{HGA+ \& LGA+} & 83.5 & 94.3 & 93.6 & 72.8 & 75.8 & 30.0 & \textbf{75.1} & -- \\
\bottomrule
\end{tabular}
\label{tab:ablation_sum}
\end{table*}

\begin{table*}[!tbp]
\centering
\tiny
\caption{Ablation study of LMSeg on BBW dataset. $\Delta$ is the mIoU difference from LMSeg (Area 1–6).}
\setlength{\tabcolsep}{4pt}\renewcommand{\arraystretch}{1.05}
\begin{tabular}{@{}lcc *{6}{ccc}@{}}
\toprule
\multirow{2}{*}{Model} & \multirow{2}{*}{$k$-nbrs.} & \multirow{2}{*}{Encoder}
& \multicolumn{3}{c}{Area1} & \multicolumn{3}{c}{Area2} & \multicolumn{3}{c}{Area3}
& \multicolumn{3}{c}{Area4} & \multicolumn{3}{c}{Area5} & \multicolumn{3}{c}{Area6} \\
\cmidrule(lr){4-6}\cmidrule(lr){7-9}\cmidrule(lr){10-12}\cmidrule(lr){13-15}\cmidrule(lr){16-18}\cmidrule(lr){19-21}
& & & F1 & mIoU & $\Delta$ & F1 & mIoU & $\Delta$ & F1 & mIoU & $\Delta$
& F1 & mIoU & $\Delta$ & F1 & mIoU & $\Delta$ & F1 & mIoU & $\Delta$ \\
\midrule
BBW-A & 3$^\ast$ & LGA+  & 58.9 & 41.8 & -19.2 & 61.7 & 44.6 & -16.2 & 75.6 & 60.7 & -9.8  & 60.6 & 43.5 & -16.5 & 59.1 & 41.9 & -16.3 & 66.9 & 50.3 & -13.9 \\
BBW-B & 32       & HGA+  & 74.3 & 59.1 & -1.9  & 74.2 & 59.0 & -1.8  & 81.9 & 69.4 & -1.1  & 73.8 & 58.4 & -1.6  & 72.1 & 56.4 & -1.8  & 77.3 & 63.1 & -1.1  \\
BBW-C & 32 (3$^\ast$) & HGA \& LGA  & 73.1 & 57.6 & -3.4  & 73.4 & 57.9 & -2.9  & 81.2 & 68.4 & -2.1  & 71.9 & 56.2 & -3.8  & 71.3 & 55.4 & -2.8  & 76.3 & 61.6 & -2.6  \\
\midrule
BBW-D & 8 (3$^\ast$)  & HGA+ \& LGA+ & 70.6 & 54.5 & -6.5  & 70.6 & 54.6 & -6.2  & 80.6 & 67.5 & -3.0  & 70.6 & 54.5 & -5.5  & 70.6 & 54.6 & -3.6  & 74.8 & 59.7 & -4.5  \\
BBW-E & 16 (3$^\ast$) & HGA+ \& LGA+ & 74.1 & 58.9 & -2.1  & 73.7 & 58.3 & -2.5  & 82.1 & 69.6 & -0.9  & 73.5 & 58.1 & -1.9  & 73.8 & 58.4 & \phantom{-}0.2 & 76.7 & 62.2 & -2.0  \\
\midrule
BBW-F & 32 (3$^\ast$) & HGA+ \& LGA+ & 75.2 & 60.3 & -0.7  & 74.4 & 59.2 & -1.6  & 82.8 & 70.7 & \phantom{-}0.2 & 73.9 & 58.6 & -1.4  & 72.0 & 56.3 & -1.9  & 77.2 & 62.9 & -1.3  \\
BBW-G & 32 (3$^\ast$) & HGA+ \& LGA+ & 76.1 & 61.4 & \phantom{-}0.4 & 75.2 & 60.3 & -0.5  & 82.9 & 70.9 & \phantom{-}0.4 & 75.6 & 60.8 & \phantom{-}0.8 & 74.6 & 59.5 & \phantom{-}1.3 & 78.1 & 64.1 & -0.1  \\
\midrule
BBW-H & 32 (3$^\ast$) & HGA+ \& LGA+ & 75.2 & 60.3 & -0.7 & 75.2 & 60.2 & -0.6 & 80.9 & 67.9 & -2.6 & 75.0 & 59.9 & -0.1 & 73.4 & 57.9 & -0.3 & 77.8 & 63.7 & -0.5 \\
BBW-I & 32 (3$^\ast$) & HGA+ \& LGA+ & 76.1 & 61.4 & +0.4 & 75.4 & 60.5 & -0.3 & 82.0 & 69.5 & -1.0 & 74.4 & 59.3 & -0.7 & 74.0 & 58.7 & +0.5 & 78.6 & 64.7 & +0.5 \\
BBW-J & 32 (3$^\ast$) & HGA+ \& LGA+ & 76.2 & 61.5 & +0.5 & 75.1 & 60.1 & -0.7 & 82.5 & 70.2 & -0.3 & 75.2 & 60.3 & +0.3 & 73.9 & 58.6 & +0.4 & 78.4 & 64.4 & +0.2 \\
\midrule
\rowcolor{lightgray}
\shortstack{\makecell[l]{LMSeg \\ (Ours-avg)}} & 16 & HGA+ \& LGA+ 
& 75.8 & 61.0 & - & 75.6 & 60.8 & - & 82.7 & 70.5 & - & 75.0 & 60.0 & - & 73.6 & 58.2 & - & 78.2 & 64.2 & - \\
\bottomrule
\end{tabular}
\label{tab:ablation_bbw}
\end{table*}

\subsection{Effectiveness of HGA+ and LGA+}  
\label{abla_hgap_lgap}

\begin{figure*}[!tbp]
    \centering
    \subfloat[Feature maps of HGA+ (top row) and LGA+ (bottom row) modules visualized on the SUM dataset.]{\includegraphics[width=0.89\textwidth]{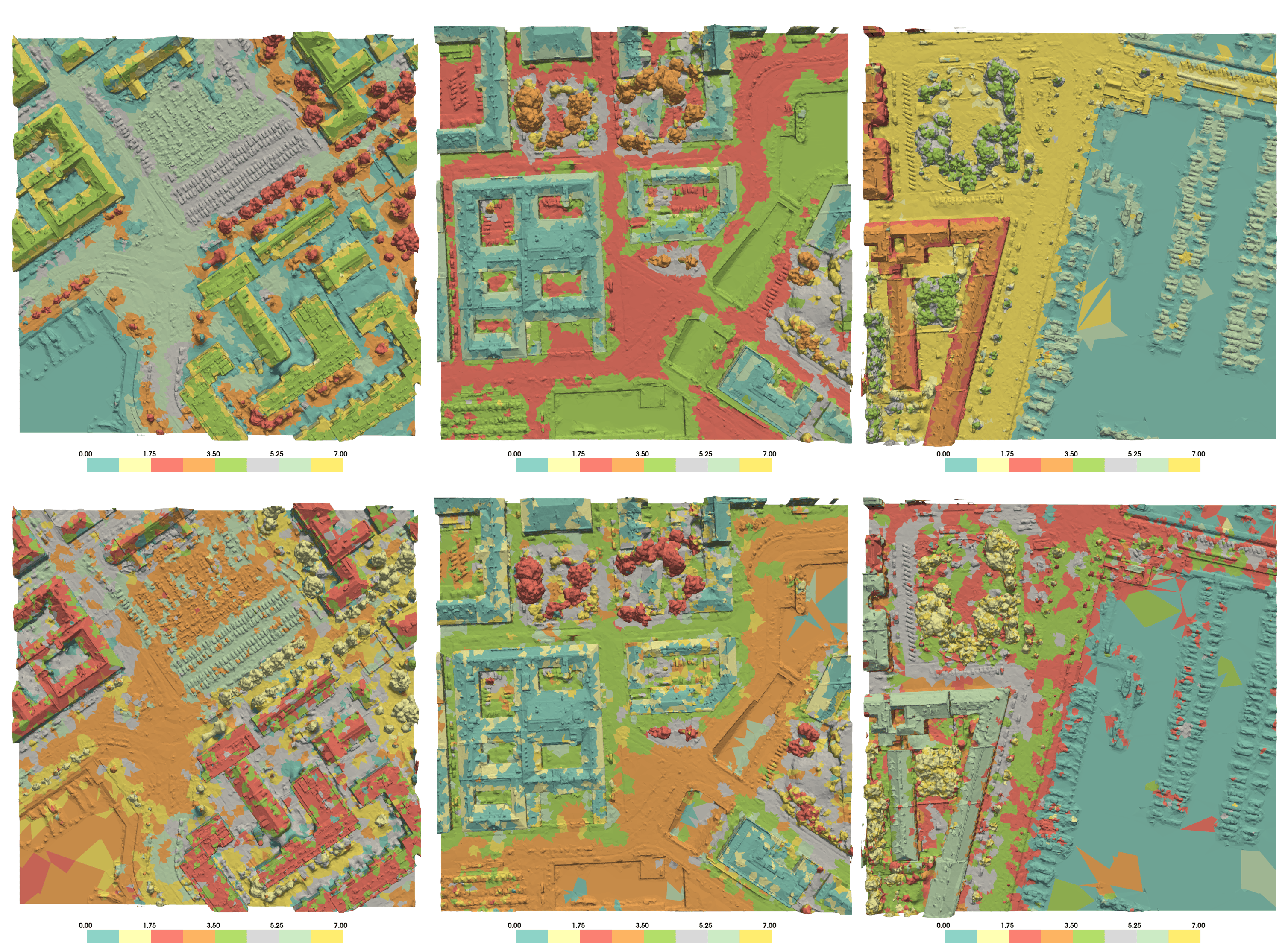}}\\
    \subfloat[Feature maps of HGA+ (top row) and LGA+ (bottom row) modules visualized on the BBW dataset.]{\includegraphics[width=0.89\textwidth]{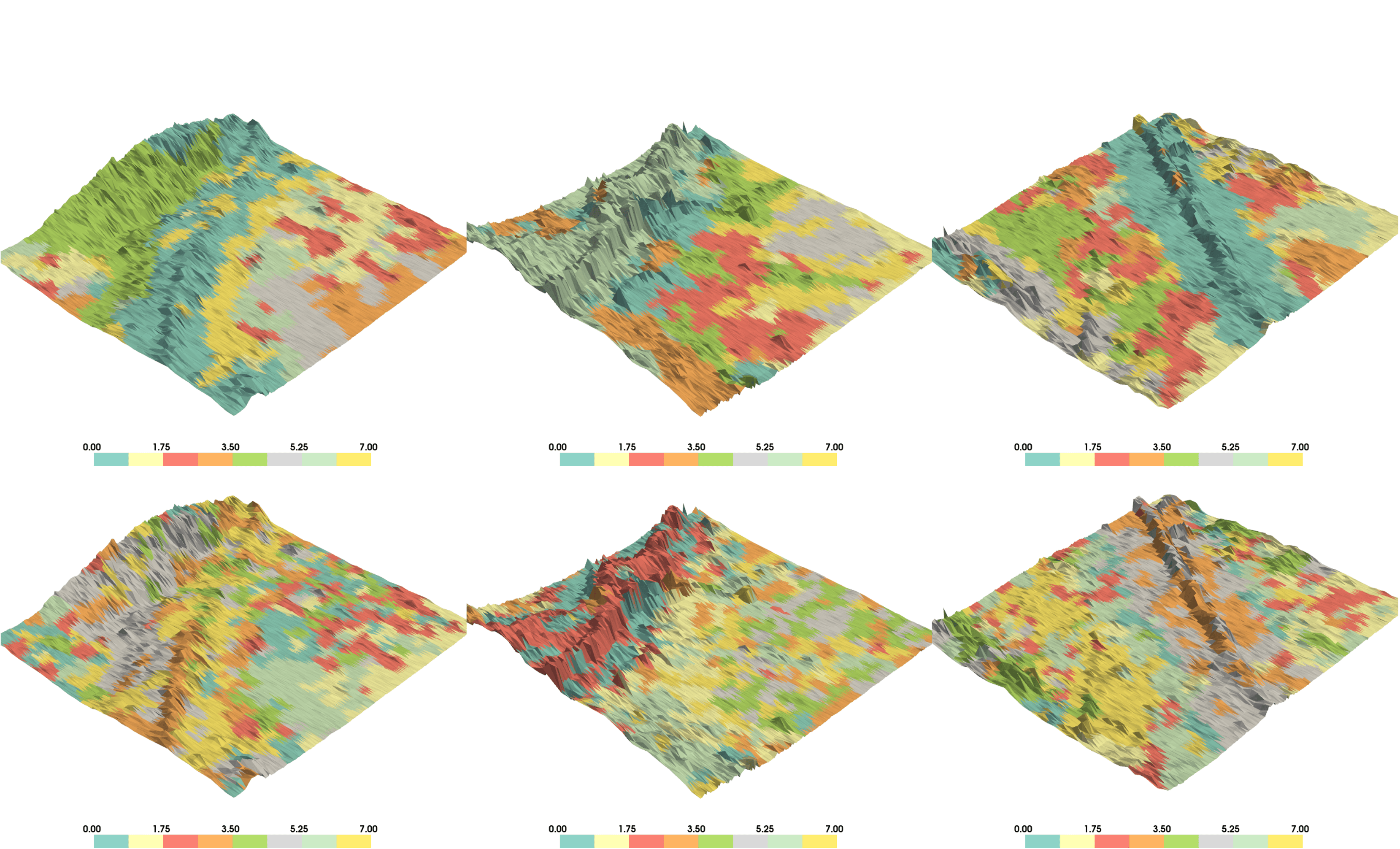}}
    \caption[]{}
    \label{fig:feature_map_p1}
\end{figure*}

\begin{figure*}[!tbp]
    \ContinuedFloat
    \centering
    \subfloat[Feature maps of HGA+ (top row) and LGA+ (bottom row) modules visualized on the H3D dataset.]{\includegraphics[width=0.89\textwidth]{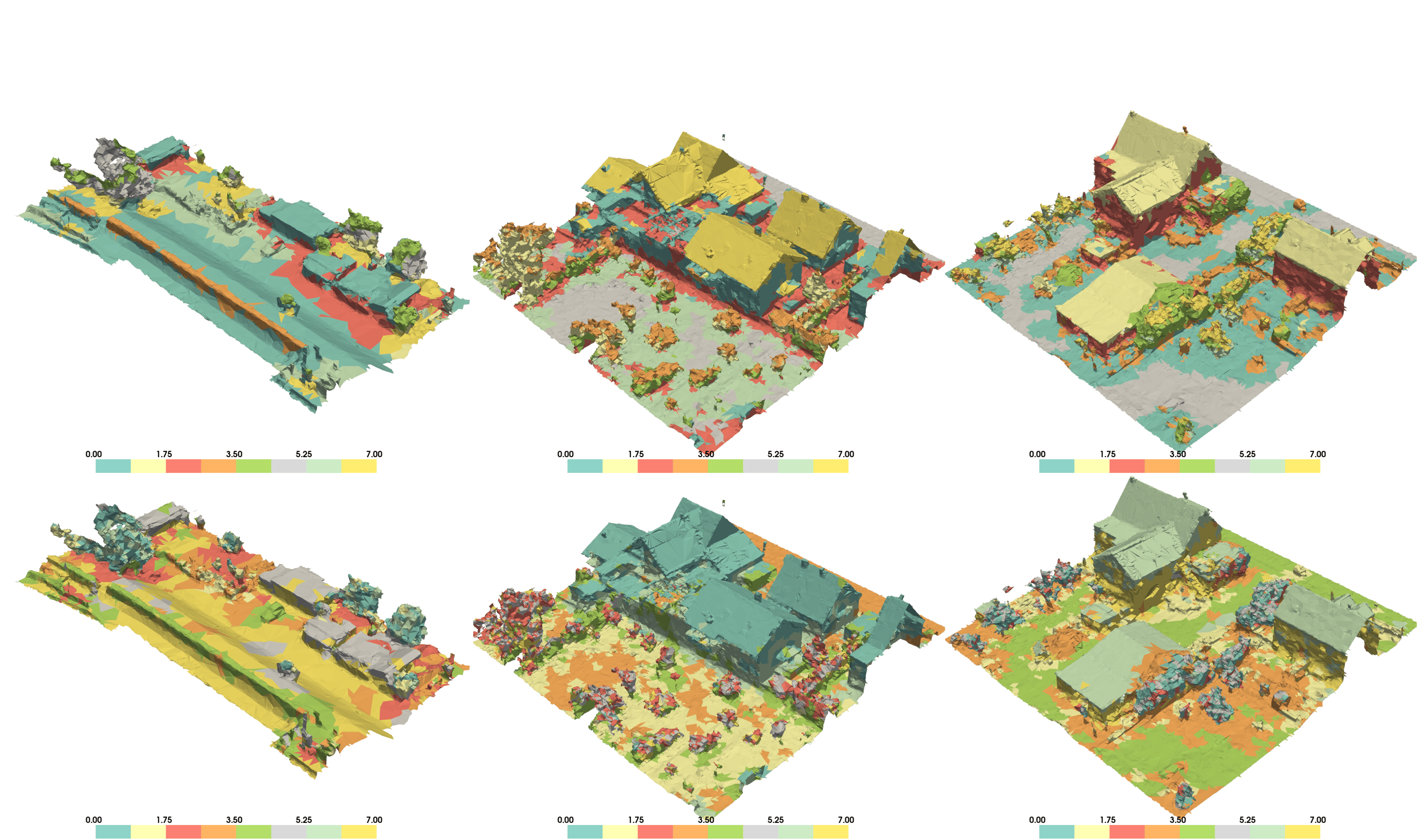}}
    \caption{Visualization of intermediate feature maps from HGA+ and LGA+ modules of LMSeg on SUM, BBW, and H3D datasets. High-dimensional embeddings are reduced with whitening, L2-normalized PCA, then clustered by K-means (8 clusters), and projected back to mesh faces. Across the datasets, HGA+ yields smooth and spatially coherent clusters that capture global geometric structure, while LGA+ produces more heterogeneous, boundary-sensitive clusters that sharpen local transitions.}
    \label{fig:feature_map_p2}
\end{figure*}

The ablation experiments highlight the complementary roles of the two aggregation modules in LMSeg. Removing HGA+ (SUM-A, BBW-A) results in substantial performance degradation, with mIoU dropping by $-16.6$\% on SUM (Table~\ref{tab:ablation_sum}) and by as much as $-19.2$\% in Area~1 of BBW (Table~\ref{tab:ablation_bbw}). These large declines confirm that hierarchical aggregation is indispensable for capturing long-range geometric context across extensive mesh regions. In contrast, removing LGA+ (SUM-B, BBW-B) produces smaller but consistent drops ($-1.8$\% on SUM and about $-1.9$\% across BBW areas). Most of these errors occur at wall and terrain boundaries, highlighting LGA+’s role in boundary refinement and fine-scale discrimination. Since edge similarity pooling is unique to LGA+, its removal also implicitly tests the effect of weakened local connectivity.

To better understand the effectiveness of HGA+ and LGA+ modules, we visualize their intermediate feature maps using PCA–KMeans clustering as display on Fig.~\ref{fig:feature_map_p1}. High-dimensional embeddings from the deepest HGA+ and LGA+ layers are reduced with PCA (whitening and L2 normalization) and clustered into 8 discrete groups with K-means clustering, with results shown as categorical color maps.

In the SUM dataset (Fig.~\ref{fig:feature_map_p1}a), the HGA+ features (top row) form broad, coherent clusters that align with large-scale regions such as vegetation clusters, built-up blocks, and open terrain, reflecting their ability to capture global structure. By contrast, the LGA+ features (bottom row) appear more fragmented, with sharply contrasted patches that follow building edges, vegetation boundaries, and terrain–water transitions, highlighting local refinement.

In the BBW dataset (Fig.~\ref{fig:feature_map_p1}b), the HGA+ features (top row) show coherent color clusters that follow terrain morphology, encoding gradual changes across hills, valleys, and ridges. The LGA+ features (bottom row) instead emphasize sharp contrasts along wall ridges and surrounding terrain, making the man-made structures stand out more distinctly against the natural landscape.

In the H3D dataset (Fig.~\ref{fig:feature_map_p2}c), the HGA+ features (top row) exhibit large, unified clusters over planar regions such as C00: Low Vegetation and C01: Impervious Surface, as well as built-up objects like C04: Roof. This demonstrates the global module’s capability to capture structural coherence even in a more semantically diverse dataset. By contrast, the LGA+ features (bottom row) are more spatially fragmented, forming smaller clusters on fine-scale categories such as C06: Shrub and C07: Tree, reflecting its sensitivity to local variations.

Together, these visualizations illustrate the complementary roles of the two modules: HGA+ enforces global structural coherence, while LGA+ sharpens boundaries and enhances local precision. Their interplay enables LMSeg to balance robustness to large-scale context with sensitivity to fine detail, supporting strong generalization across urban (SUM, H3D) and archaeological (BBW) landscapes.

\subsection{Learnable Aggregation for GA+}

\begin{figure*}[!tbp]
    \centering
    \subfloat[SUM dataset.]{\includegraphics[width=0.78\textwidth]{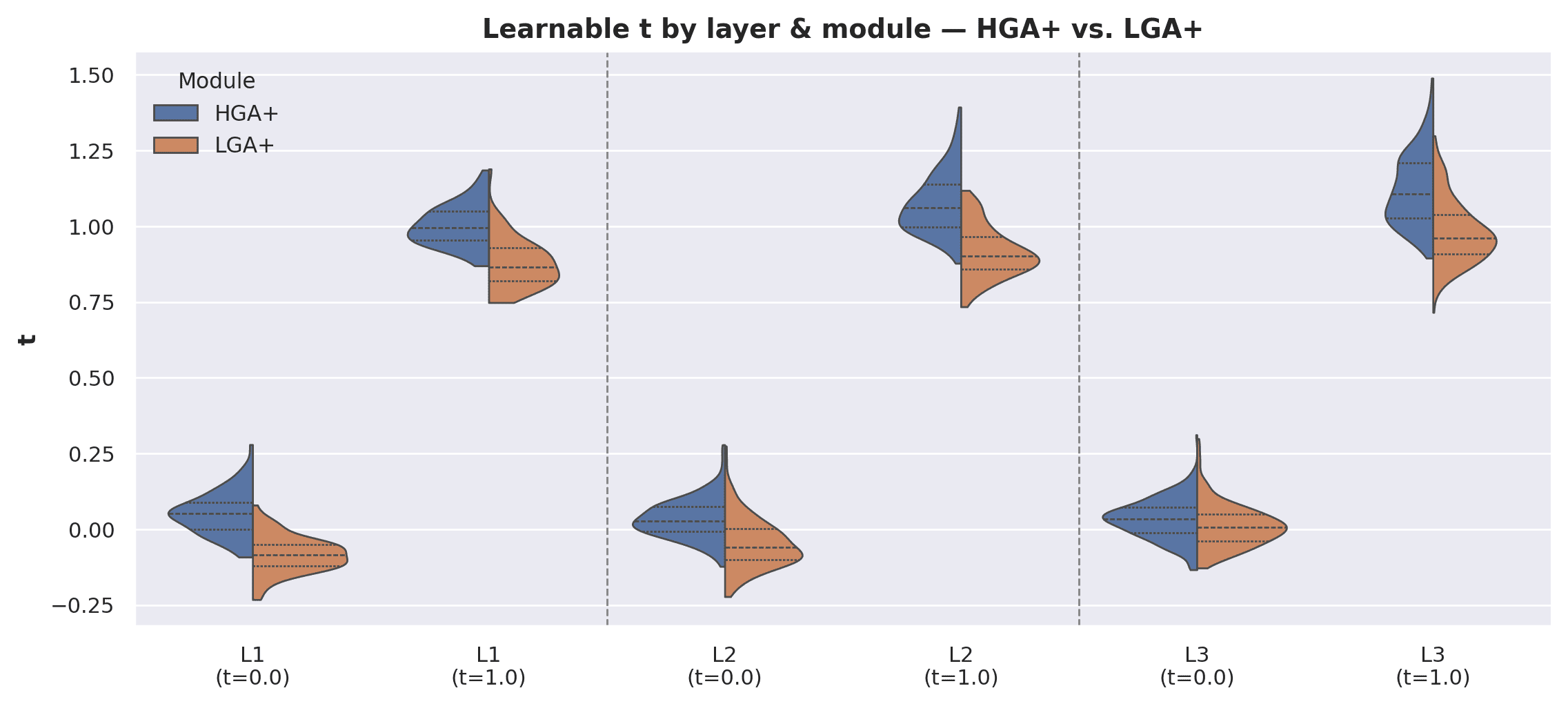}} \\
    \subfloat[BBW dataset.]{\includegraphics[width=0.78\textwidth]{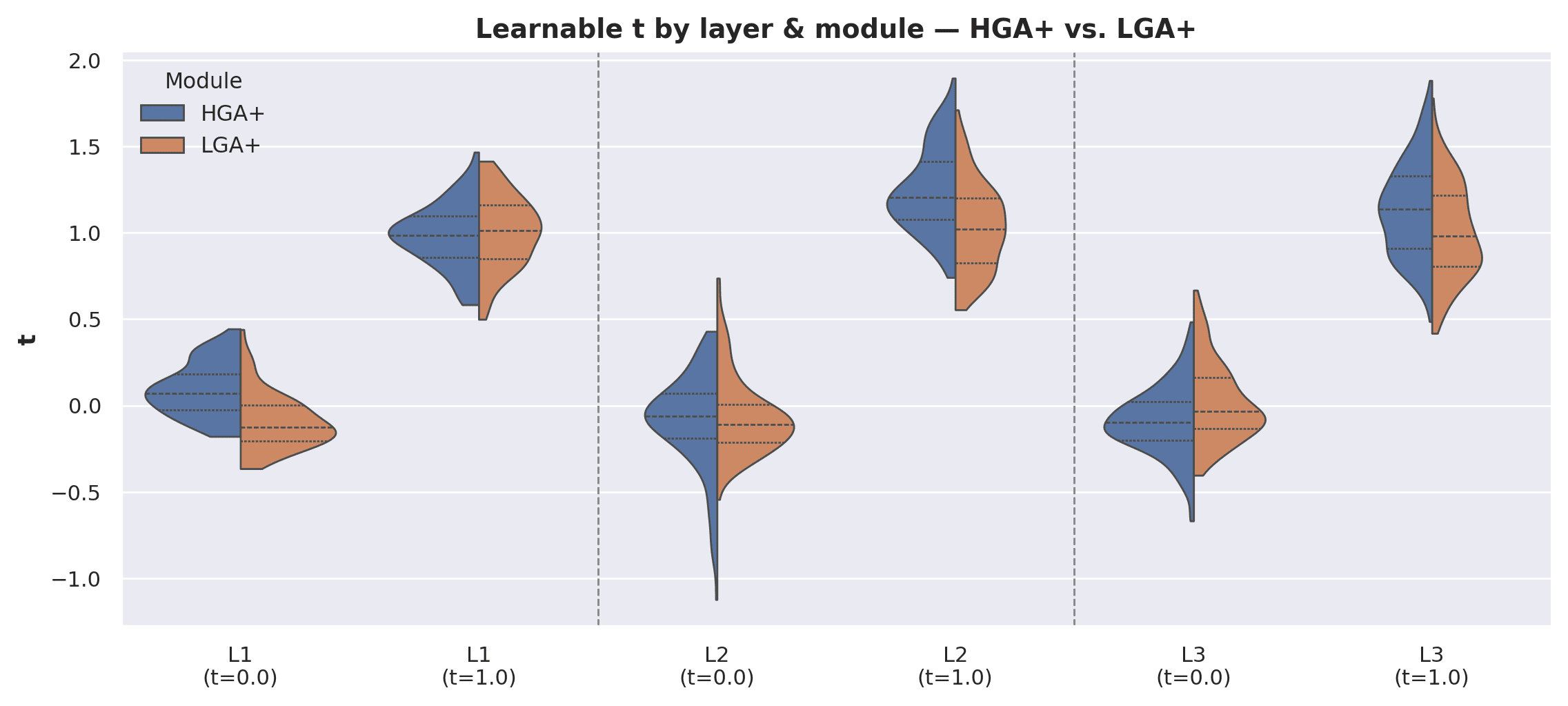}} \\
    \subfloat[H3D dataset.]{\includegraphics[width=0.78\textwidth]{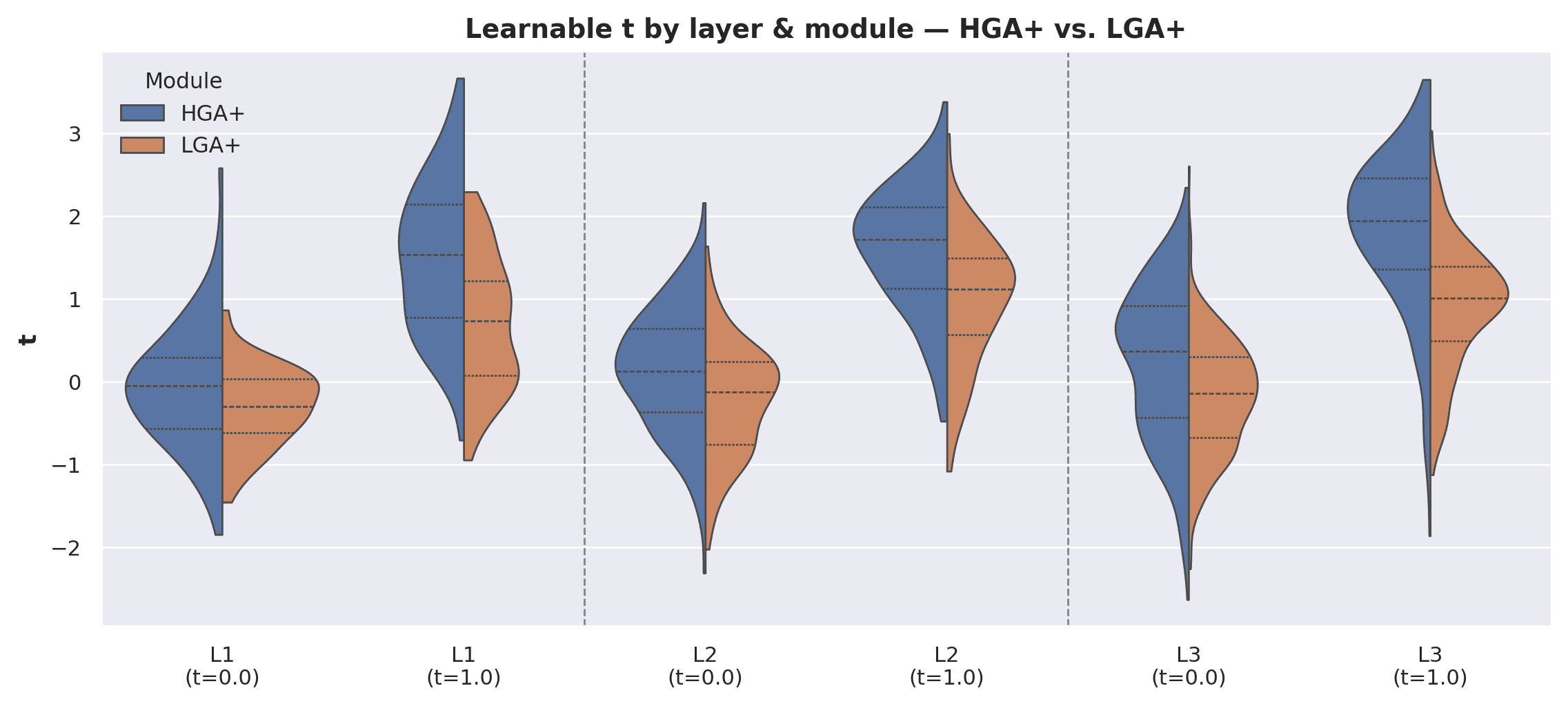}}   
    \caption{Distribution of the learnable softmax temperature parameter $\textbf{t}$ in HGA+ and LGA+ modules on the SUM, BBW and H3D datasets. Each violin plot shows the spread of t values across layers (1–3) for hierarchical aggregation (HGA+, blue) and local aggregation (LGA+, orange) modules. Two groups at each layer corresponds to the average-like aggregation (t=0.0) and the max-like aggregation (t=1.0) operators. Overall, HGA+ tends to concentrate toward sharper, max-like aggregation in deeper layers, while LGA+ maintains a broader spread between averaging and maximization, reflecting its role in balancing local stability and fine-scale sensitivity.}
    \label{fig:softmax_aggr}
\end{figure*}

The GA+ operator provides further gains by replacing fixed aggregation with a learnable scheme. When the learnable aggregation scheme is replaced with fixed aggregation in GA+ (SUM-C, BBW-C), mIoU decreases by $-6.1$\% on SUM and by $-2$ to $-3.8$\% across BBW areas. These declines show that fixed operators (e.g., mean or max) cannot capture the varying importance of neighbors in different geometric contexts, whereas GA+ improves robustness by adaptively interpolating between them through a learnable inverse temperature parameter.

The distributions of this parameter $t$ in HGA+ and LGA+ modules are visualized in Figure~\ref{fig:softmax_aggr} for the SUM, BBW, and H3D datasets. The violin plots show how $t$ values are spread across layers 1–3 of LMSeg for both average-like ($t=0.0$) and max-like ($t=1.0$) aggregation operators (Eq.~\ref{eq:aggr_softmax_add}). In HGA+, $t$ tends to cluster near the max-based regime, especially in deeper layers, indicating that hierarchical aggregation increasingly emphasizes strong neighbor signals for capturing long-range structural context. In contrast, LGA+ exhibits a broader spread of $t$, often interpolating between mean- and max-based regimes. This reflects the local module’s need to balance stability (via averaging) with sensitivity to fine-scale boundary variations (via maximization). The consistent distribution patterns observed across all three datasets validate the complementary functions of the two modules: HGA+ progressively sharpens global structure, while LGA+ flexibly adapts to local heterogeneity. 

These results not only validate the learnable design of GA+ but also explain why fixed aggregation operators fail to generalize across diverse contexts. In particular, HGA+ gravitates toward sharper, max-like aggregation in deeper layers, whereas LGA+ maintains a broader balance between mean- and max-based regimes, mirroring their respective global and local roles.

\subsection{Mesh Feature Ablation}
Feature ablations highlight the complementary contributions of geometry and texture. Excluding RGB (SUM-F, BBW-F) produces the steepest declines, with mIoU dropping $-7.6$\% on SUM and up to $-1.9$\% on BBW. By comparison, excluding normals (SUM-G, BBW-G) yields much smaller decreases ($-1.2$\% on SUM, $-0.1$\% on BBW). This indicates that RGB is the dominant cue in these datasets. On SUM, RGB provides clear class distinctions between vegetation, buildings, and water, whereas normals are often redundant with coordinates. On BBW, normals are degraded by vegetation occlusion and rough terrain, while RGB still provides subtle reflectance cues that help distinguish walls from natural surroundings. Taken together, these results suggest that LMSeg benefits from both modalities but relies more heavily on texture cues when geometry is noisy or ambiguous.

\subsection{Sensitivity Analysis of Model}
We evaluated LMSeg under sensitive parameter settings to assess robustness against variations in hierarchical neighborhood size (SUM-/BBW-D and SUM-/BBW-E), feature dimensionality (SUM-/BBW-H), and learning rate (SUM-/BBW-I and SUM-/BBW-J) (Tab.~\ref{tab:ablation_sum}, Tab.~\ref{tab:ablation_bbw}). These experiments complement the main ablations by probing whether performance remains stable under more extreme parameter choices.

\textbf{Neighborhood size.} Reducing the HGA+ neighborhood from $k=32$ to $k=8$ (SUM-D, BBW-D) leads to clear declines ($-2.9$\% on SUM, $-6.5$\% on BBW Area~1), showing that overly small neighborhoods limit contextual aggregation and hinder boundary reconstruction. Using $k=16$ (SUM-E, BBW-E) reduces this gap, with moderate drops ($-1.6$\% on SUM, $-2.1$\% on BBW Area~1). These results indicate that the model benefits from sufficiently large neighborhoods to capture global context, while remaining moderately robust to reductions.

\textbf{Feature dimensionality.} Reducing the initial feature space from $D=72$ to $D=36$ (SUM-H, BBW-H) causes moderate performance drops ($-3.6$\% on SUM, $-0.7$\% on BBW Area~1). The sharper decline on SUM suggests that complex, heterogeneous urban scenes demand higher representational capacity, whereas BBW is less sensitive. Retaining $D=72$ thus offers a balanced trade-off between expressiveness and compactness.

\textbf{Learning rate.} Adjusting $lr$ upward to $0.005$ (SUM-I, BBW-I) or downward to $0.0005$ (SUM-J, BBW-J) results in only small changes ($\pm 0.9$\% on SUM, $\pm 0.5$\% on BBW). This demonstrates that LMSeg converges reliably over a reasonable range of optimization settings, with low sensitivity to learning rate variation.

Overall, LMSeg is most affected by reductions in neighborhood size and feature dimensionality, which directly constrain its ability to encode hierarchical geometric context. By contrast, training remains stable across a broad learning rate range. These findings reinforce that LMSeg delivers strong segmentation accuracy without requiring delicate hyperparameter tuning, underscoring its robustness for large-scale mesh segmentation.

\section{Conclusions}
\label{conclusion}

We presented LMSeg, a deep graph message-passing network for semantic segmentation of large-scale 3D landscape meshes, and introduced the BudjBim Wall (BBW) dataset for mapping historic dry-stone walls in vegetated terrain. LMSeg addresses three persistent challenges in mesh segmentation: limited scalability, lack of end-to-end trainability, and difficulty in capturing small or irregular objects. By leveraging a barycentric dual graph representation and the Geometry Aggregation+ (GA+) module, LMSeg jointly captures hierarchical and fine-grained geometric features, enabling accurate segmentation across both urban and natural environments. Experiments on the SUM, H3D, and BBW benchmarks demonstrate strong performance (i.e., 75.1\% mIoU on SUM, 78.4\% O.A. on H3D, and 62.4\% mIoU on BBW) with a compact 2.4M-parameter model.

LMSeg offers advantages in small-object segmentation, multi-scale feature learning, and end-to-end trainability, but some limitations remain. Its reliance on mesh-only features may reduce robustness in scenarios where complementary modalities (e.g., LiDAR intensity, aerial imagery) provide essential context. In addition, random sub-sampling in HGA+ modules can discard structurally important details in highly complex regions. While LMSeg demonstrates consistent performance across three distinct datasets, a direct cross-domain transfer evaluation (e.g., training on SUM and testing on H3D) has not yet been conducted. Robustness across unseen domains therefore remains an open challenge and an important direction for future research. Future work should also explore multimodal integration and geodesically informed or learnable pooling strategies to better preserve topology, as well as extend evaluations to diverse landscapes and geographic regions to assess broader generalization.

Overall, LMSeg and the BBW dataset provide a technically sound and practically relevant framework for advancing semantic segmentation of large-scale 3D meshes, with applications in cultural heritage documentation, environmental monitoring, and urban modeling.

\section*{CRediT authorship contribution statement}
\textbf{Zexian Huang}: Conceptualization, Writing – original draft, Writing – review \& editing, Software, Methodology, Investigation. \textbf{Kourosh Khoshelham}: Writing – review \& editing, Supervision. \textbf{Martin Tomko}: Writing – review \& editing, Supervision.

\section*{Acknowledgements}
\label{acknowledgements}
We thank the Gunditj Mirring Traditional Owners Corporation (GMTOC) for providing the research data and overseeing its use in this study. We also acknowledge Mashnoon Islam for his dedicated work on data annotation. Their invaluable cooperation, support, and meticulous efforts have significantly contributed to the quality of this research. Permission has been obtained from all individuals and organizations mentioned in this acknowledgment.

\section*{Declaration of Competing Interest}
The authors declare that they have no known competing financial interests or personal relationships that could have appeared to influence the work reported in this paper.

\section*{Declaration of Funding.} No funding or sponsorship was obtained for this study.

\section*{Data availability}
\label{availability}
The source code and data supporting the findings of this study are available at: \url{https://github.com/zexhuang/LMSeg}.

\appendix


\bibliographystyle{elsarticle-harv} 
\bibliography{cas-refs}





\end{document}